\documentclass[letterpaper,twocolumn,10pt]{article}
\usepackage{usenix-2020-09}

\usepackage{array}      
\usepackage{siunitx}    

\newcolumntype{S}{}
\usepackage{amsthm}
\usepackage{amsmath}
\usepackage{amsfonts}
\usepackage{graphicx}
\usepackage{threeparttable}
\usepackage{xcolor}
\usepackage{xspace}
\usepackage{soul}
\usepackage[normalem]{ulem}
\usepackage{algorithm}
\usepackage{enumitem}
\usepackage{pifont}
\usepackage{multirow}
\usepackage{tikz}
\usepackage{textcomp}
\usepackage{booktabs}
\usepackage{subcaption}
\usepackage{float}
\usepackage[utf8]{inputenc}
\usepackage{textgreek}
\usepackage{comment}
\usepackage{makecell}
\usepackage{booktabs}
\usepackage{graphicx}
\usepackage{multirow}
\usepackage{algpseudocode}
\definecolor{commentcolor}{RGB}{60, 128, 49}
\algrenewcommand{\algorithmiccomment}[1]{\textcolor{commentcolor}{\ \(\triangleright\) #1}}

\newcommand{\sysname}{LiteCache\xspace}
\newcommand{\algoname}{QSAC\xspace}
\newcommand{\Topk}{Top-$k$\xspace}
\newcommand{\topk}{top-$k$\xspace}

\newcommand{\para}[1]{\noindent \textbf{#1 }}

\usepackage{CJKutf8}

\definecolor{darkgreen}{rgb}{0.078,0.667,0.016}

\newcommand{\zwc}[1]{\textcolor{magenta}{[\begin{CJK}{UTF8}{gbsn}
ZW-Comment:~#1\end{CJK}]}}

\usepackage[compact]{titlesec}
\titlespacing*{\subsection}{0pt}{1ex}{0.5ex}
\titlespacing*{\subsubsection}{0pt}{1ex}{0.5ex}
\titlespacing{\section}{0pt}{1ex}{0.8ex}

\setlist[itemize]{noitemsep, topsep=0pt}

\begin{document}

\date{}

\title{\Large \bf \sysname: A Query Similarity-Driven, GPU-Centric KVCache Subsystem for Efficient LLM Inference}

\author{
{\rm Jiawei Yi} \\ University of Science and Technology of China
\and
{\rm Ping Gong} \\ University of Science and Technology of China
\and
{\rm Youhui Bai} \\ University of Science and Technology of China
\and
{\rm Zewen Jin} \\ University of Science and Technology of China
\and
{\rm Shengnan Wang} \\ Independent Researcher
\and
{\rm Jiaqi Ruan} \\ University of Science and Technology of China
\and
{\rm Jia He} \\ University of Science and Technology of China
\and
{\rm Jiaan Zhu} \\ University of Science and Technology of China
\and
{\rm Pnegcheng Wang} \\ Huawei Technologies Co., Ltd
\and
{\rm Haibo Wang} \\ Huawei Technologies Co., Ltd
\and
{\rm Weiguang Wang} \\ Huawei Technologies Co., Ltd
\and
{\rm Xia Zhu} \\ Huawei Technologies Co., Ltd
\and
{\rm Cheng Li} \\ University of Science and Technology of China
} 


\maketitle

\begin{abstract}

\noindent During LLM inference, KVCache memory usage grows linearly with sequence length and batch size and often exceeds GPU capacity. Recent proposals offload KV states to host memory and reduce transfers using \topk attention. But their CPU-centric management of the on-GPU cache and CPU-GPU data movement incurs high overhead and fragments the bulk GPU execution that CUDA Graph relies on.


To close this gap, we observe that adjacent queries within the same attention head exhibit strong directional similarity and retrieve highly overlapping \topk KV states. This insight enables a simple head granularity cache algorithm, \algoname, in which each head reuses its previously cached KV states whenever the current query is sufficiently similar to the prior one. \algoname further simplifies cache management primitives and cuts CPU involvement almost entirely. We develop \sysname, a KVCache subsystem that incorporates \algoname. \sysname introduces a GPU-centric synchronization controller and speculative sparse prefetching, enabling fully overlapped data movement and computation. These mechanisms produce a stable and predictable execution pattern that remains compatible with the bulk execution mode required by CUDA Graphs.

Evaluation on two widely-used LLMs indicates that \sysname achieves comparable accuracy to baselines, while sharply minimizing CPU overhead, fully utilizing PCIe bandwidth, thus improving decoding throughput by 10.7-224.2\% on both H100 and A40 GPUs and easily supporting sequence lengths beyond 1M. 
We opensource \sysname at~\cite{LiteCache2025}. 

\end{abstract}

\section{Introduction}
\label{sect:intro}

\noindent In recent years, generative large language models (LLMs)~\cite{vaswani2017attention} have been widely adopted in real-world applications, such as natural language processing~\cite{qin2024llmfornlp}, code generation~\cite{jiang2024llmforcoding}, and multimodal tasks~\cite{yin2024llmformultimodal}. Modern LLMs adopt the Transformer architecture and maintain a key–value cache (KVCache) during autoregressive decoding to store past attention states. 

However, KVCache memory footprint scales linearly with sequence length and batch size, often exceeding the GPU HBM capacity. For example, a 512K token input to Qwen2.5–14B produces a KVCache of 93.75 GB, more than 3$\times$ the model’s parameter size. As models and context windows continue to grow, the tension between limited GPU memory and expanding KVCache demands increasingly constrains the scalability of LLM inference.

To overcome this limitation, recent KVCache subsystems offload KV data to host memory and dynamically fetch only the portions needed for each decoding step~\cite{sheng2023flexgen,lee2024infinigen,zhang2025pqcache,chen2025retroinfer,xiao2024infllm}. To further reduce KV data transferred over PCI-e, they incorporate \topk attention, which exploits the sparsity of attention to select only the most relevant KV states for each query. Building on this mechanism, advanced subsystems such as RetroInfer~\cite{chen2025retroinfer} and PQCache~\cite{zhang2025pqcache} further maintain a on-GPU cache managed by general purpose replacement algorithms (e.g., LRU~\cite{o1993lru} and LFU~\cite{robinson1990data}). This two-tier design alleviates GPU memory pressure and reduces data movement compared to naive offloading. However, these subsystems still fall far short of the performance achieved by a full attention baseline, due to the following two problems.

First, CPU-side cache management has become a major scalability barrier. Each decoding step triggers CPU-driven lookup, data transfer planning, metadata updates, and on-GPU cache replacement. These operations grow costly with longer contexts, larger batches, and larger cache sizes, leading to substantial CPU overhead and GPU stalls. Empirically, we observe that cache-related CPU time can reach 68\% of GPU compute time on an H100~\cite{H100Whitepaper2025} (Table~\ref{tab:handling_overhead}), causing GPU to idle while waiting for cache metadata and KV blocks to arrive.

Second, CPU–GPU coordination prevents the use of CUDA Graphs~\cite{cuda-graph}, a key optimization for modern inference. High-end GPUs such as NVIDIA H100 offer massive compute throughput, yet kernel launch overhead now dominates execution, reaching up to 60\% of end-to-end latency in our measurements. CUDA Graphs can greatly reduce this overhead by capturing a stable GPU execution pattern; however, existing KVCache subsystems introduce dynamic, fine-grained CPU–GPU synchronizations and memory transfers that fragment the GPU pipeline and break graph capture. As a result, state-of-the-art KVCache systems cannot benefit from CUDA Graphs, widening the performance gap between full attention and offloading-based approaches. 
Bridging this gap requires rethinking KVCache subsystem design from the ground up: reducing cache management to a lightweight, predictable, GPU-centric primitive while retaining the flexibility to serve \topk sparse attention efficiently.


To bridge this gap, we present \sysname, a KVCache subsystem that uses lightweight, predictable, GPU centric primitives while retaining the flexibility to serve \topk sparse attention efficiently. At the core of \sysname is the \algoname algorithm, motivated by our key observation that adjacent queries within the same attention head exhibit strong directional similarity and retrieve highly overlapping top k KV states. This stable intra-head locality allows cache decisions at the granularity of heads rather than blocks. \algoname eliminates list based metadata, stores only the previous query per head, and performs lookup with a single GPU kernel that checks similarity. When similarity is high, the head reuses its cached KV states with no block level metadata updates or merge operations, greatly simplifying cache management and reducing CPU involvement.

Building on this lightweight algorithm, \sysname introduces a set of system level mechanisms that address the CPU–GPU coordination challenges of existing KVCache subsystems. \sysname leverages the direction of each query to predict which \topk KV states are likely to be needed and issues speculative sparse prefetching from CPU memory. These transfers proceed concurrently with GPU execution, allowing data movement and computation to overlap. To maintain a GPU driven pipeline and avoid frequent CPU intervention, \sysname adds a GPU side synchronization controller that polls the completion state of data transfers entirely on the GPU. This removes the fine grained CPU orchestrations that fragment execution and break CUDA Graph replay. Together, these mechanisms allow \sysname to execute sparse attention and KVCache offloading with a smooth, static, and bulk GPU execution pattern, enabling high throughput decoding on modern accelerators.

We implemented \sysname atop FlashInfer~\cite{ye2025flashinfer} and Transformers~\cite{wolf-etal-2020-transformers}, and open sourced the system at~\cite{LiteCache2025}. We conduct an extensive evaluation using two major LLM families (Llama3-8B~\cite{gradientlongcontextllama3} and Qwen2.5-14B~\cite{qwen2.5-1m}) across two GPU platforms (NVIDIA A40 and H100), covering a range of sequence lengths from 4K to 1M. 
Compared to state-of-the-art KVCache subsystems, \sysname delivers substantial performance gains, improving decoding throughput by 67.0–224.2\% on H100 and 10.7–60.5\% on A40, while achieving accuracy comparable to full attention on both LongBench~\cite{bai-etal-2024-longbench} and RULER~\cite{hsieh2024ruler}.
This shows that combining similarity-aware caching, GPU-centric execution, and model-driven design can overcome long-standing KV-cache bottlenecks and enable scalable long-context inference on modern accelerators.

\section{Background and Motivation}
\label{Sec:background}

\subsection{LLM Decoding Inefficiency}
\label{Sec:background:llm_inference}
Modern large language models (LLMs) are built upon the Transformer architecture, whose attention mechanism enables the model to capture contextual relationships across input tokens~\cite{vaswani2017attention,gradientlongcontextllama3,llama3.1,qwen2.5-1m}. Among the many variants of attention, multi-head attention (MHA)~\cite{vaswani2017attention} is the most widely adopted. MHA employs $h$ parallel attention heads that jointly process the input. Given input hidden states $\mathbf{X} \in \mathbb{R}^{n \times d}$, where $n$ is the sequence length and $d$ is the hidden dimension, each head $i$ uses its own projection matrices $\mathbf{W}^i_Q, \mathbf{W}^i_K, \mathbf{W}^i_V \in \mathbb{R}^{d \times d_k}$ to obtain the Query, Key, and Value representations $\mathbf{Q}_i, \mathbf{K}_i, \mathbf{V}_i \in \mathbb{R}^{n \times d_k}$ with $d_k = d / h$. The attention for head $i$ is then computed as $\mathbf{O}_i = \text{softmax}\left( \mathbf{Q}_i \mathbf{K}_i^\top/\sqrt{d_k} \right) \mathbf{V}_i$. In the end, the outputs of all heads are concatenated and projected through $\mathbf{W}_O \in \mathbb{R}^{d \times d}$ to form the final attention output. Furthermore, recent LLMs adopt grouped query attention (GQA)~\cite{gradientlongcontextllama3,llama3.1,qwen2.5-1m}, which reduces memory and computation by allowing multiple query heads to share a single pair of KV projections.

Popular transformer-based models, such as GPT~\cite{brown2020languagemodelsfewshotlearners} and
Llama~\cite{touvron2023llama}, employ a decoder-only structure. Each inference request is logically divided into two stages: the \textit{prefill}
stage and the \textit{decoding} stage. The prefill phase processes the full prompt in parallel to generate the first token, while storing intermediate results of computed keys and values, referred to as the \textit{KVCache}. 
The decoding stage then uses KVCache to autoregressively generate new tokens, processing only one token at a time due to the constraints of autoregressive generation.


LLM inference faces significant decoding bottlenecks stemming from the KVCache. First, the process is memory constrained~\cite{ribar2023sparq} because the storage required for the KVCache grows linearly with both batch size and input sequence length. As a result, its memory footprint can easily exceed the high bandwidth memory (HBM) capacity of even high-end GPUs such as H100. For instance, in Qwen2.5-14B-Instruct-1M~\cite{qwen2.5-1m}, a 512K token input produces a KVCache of 93.75 GB, which is 3.3$\times$ larger than the model’s 28 GB parameters. Second, during each decoding step, the model must retrieve the full set of previously stored Key and Value vectors (denoted by KV states), leading to substantial data movement overheads that also scale with sequence length and batch size. These bottlenecks are further exacerbated by emerging applications that naturally involve long contexts, such as multi-round dialogue~\cite{gao2024multiturnconversation}, document summarization~\cite{koh2022documentsum}, and code understanding~\cite{bairi2024codeplan}. Meanwhile, modern inference systems increasingly favor large batch sizes to maximize throughput~\cite{kwon2023efficientmemorymanagementlarge}, compounding the memory and bandwidth pressure imposed by the KVCache.

\subsection{KVCache Subsystems}
\label{Sec:background:kvoffloading}

KVCache subsystems is a critical component of modern LLM inference and have attracted growing research attention~\cite{kwon2023efficientmemorymanagementlarge}. To mitigate GPU memory pressure, various systems offload KVCache from GPU HBM to the much larger host memory, thereby supporting longer contexts and larger batches than GPU capacity alone would allow~\cite{lee2024infinigen,sheng2023flexgen}. In parallel, another line of work aims to reduce the bandwidth cost of loading KVCache during decoding. \Topk attention exploits the intrinsic sparsity of attention~\cite{gupta2021topkattn,ribar2023sparq} by retrieving only the most relevant KV states, achieving near lossless model quality while significantly lowering data transfer volume. Advanced algorithms such as Quest~\cite{tang2024quest}, HATA~\cite{gong2025hata}, and Loki~\cite{singhania2024loki} focus on efficiently identifying the critical KV states. To this end, they generate compact \textit{retrieval metadata} derived from the KCache, such as token block descriptors~\cite{tang2024quest}, hash codes~\cite{gong2025hata}, and critical channels~\cite{singhania2024loki}, to facilitate \topk KV selection. 

\begin{figure}[!t] 
    \centering
    \includegraphics[width=\columnwidth]{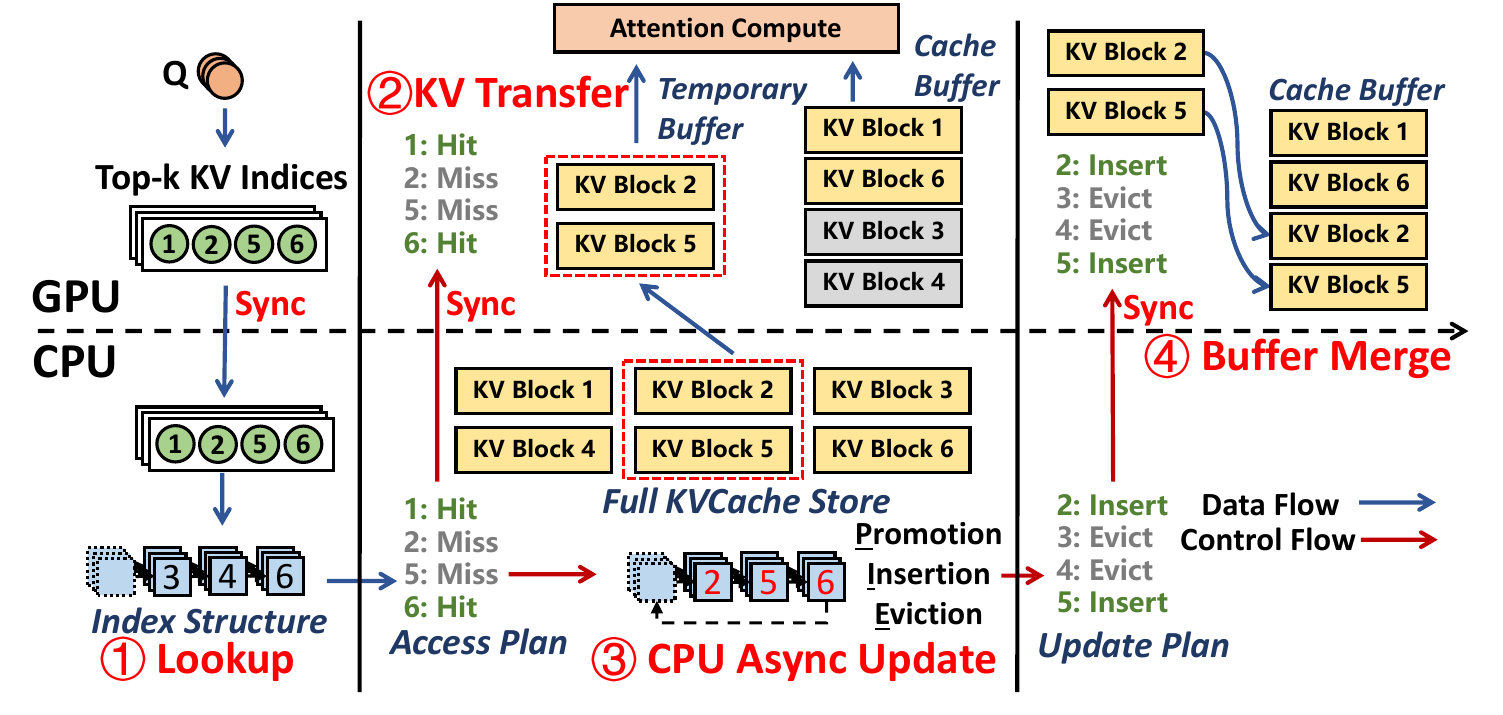}
    \caption{Overview of RetroInfer~\cite{chen2025retroinfer}'s KVCache subsystem design. Note that PQCache~\cite{zhang2025pqcache} follows a similar cache handling workflow but performs cache lookup on GPU.}
    \label{fig:background:cache_overview}
\end{figure}

More recent KVCache subsystems integrate both offloading and \topk attention, and additionally use spare GPU memory as a cache for frequently accessed KV data. Representative systems such as RetroInfer~\cite{chen2025retroinfer} and PQCache~\cite{zhang2025pqcache} adopt a unified two tier architecture~\cite{sheng2023flexgen,lee2024infinigen,zhang2025pqcache,chen2025retroinfer,xiao2024infllm}. As illustrated in Figure~\ref{fig:background:cache_overview}, the full KVCache resides in CPU-side \textit{KVCache Store}, while GPU maintains a \textit{Cache Buffer} that holds the hottest KV blocks. KV data is managed at block granularity, which typically corresponds to tens to hundreds of tokens.


For managing on-GPU cached data, these systems employ standard cache replacement policies, including LRU~\cite{o1993lru}, ARC~\cite{megiddo2003arc}, and LFU~\cite{robinson1990data}. In detail, each KV block is associated with a list structure used for replacement decisions (insertion, promotion and eviction), and an index structure (hash map) that locates blocks in CPU or GPU memory.



\noindent\textbf{Cache-driven inference workflow.} 
For every query, the GPU first computes the \topk KV indices required for the current attention operation based on the retrieval metadata produced by the \topk attention algorithm. 
These indices are then passed to the CPU, which performs a block-level \textit{CPU lookup} in the index structure to determine their locations (\ding{172}). 
The lookup generates an access plan that specifies the location of each required block. Blocks found in the GPU cache buffer are marked as hits (e.g., blocks 1 and 6), while those residing in host memory are marked as misses (e.g., blocks 2 and 5). 
For misses, the CPU gathers the corresponding KV blocks from the full KVCache store and \textit{transfers} them to the GPU over PCIe (\ding{173}). 
The GPU places the incoming data into a \textit{temporary buffer}, which is merged with cache-hit data and consumed by the attention computation. 
Meanwhile, the CPU-side workflow \textit{updates} the list structure using the hit and miss information recorded in the access plan (\ding{174}). 
At step-\ding{175}, the updated list structure then produces a \textit{buffer update plan} that instructs the GPU to merge the KV data in the temporary buffer into the cache buffer (e.g., blocks 2 and 5)  and evict out-of-dated data (e.g., blocks 3 and 4).

\subsection{Problems and Motivations}
\label{Sec:background:motivation}


The performance of KVCache subsystems plays a vital role in determining the overall efficiency of LLM inference, particularly decoding throughput and latency. To this end, we evaluate the per-step decoding latency of representative KVCache subsystems, including PQCache~\cite{zhang2025pqcache} and RetroInfer~\cite{chen2025retroinfer}, compared against a strong full attention baseline (denoted FullAttn) where the entire KVCache is stored in GPU HBM and \topk attention is disabled. We additionally evaluate a variant enhanced with CUDA Graphs (denoted FullAttn+CuGraph), an important optimization that reduces kernel launch overhead on high end GPUs~\cite{cuda-graph}. Note that current KVCache subsystems do not yet integrate CUDA Graphs, and incorporation is nontrivial; we discuss these challenges in Section~\ref{sect:bg:sync}. All experiments benchmark Llama3-8B~\cite{gradientlongcontextllama3} on NVIDIA H100 and A40 GPUs across varying sequence lengths.


\begin{figure}[!t] 
    \centering
    \includegraphics[width=0.995\columnwidth]{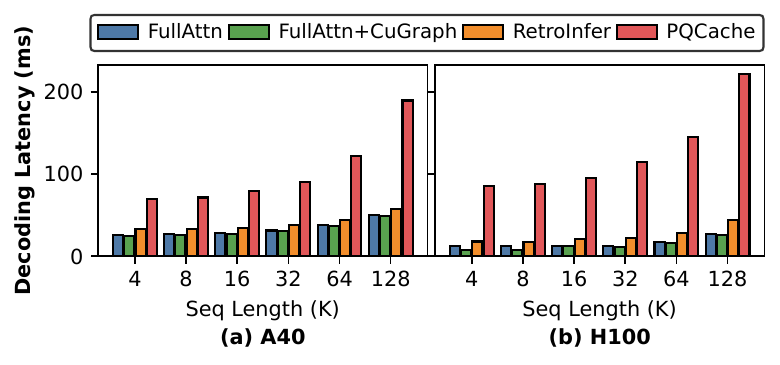}
    \caption{Per-step decoding latency comparison between on-GPU caching systems and full-attention baselines with varied sequence lengths.}
    \label{fig:background:n2n_latency}
\end{figure}

\begin{table}[!t]
    \centering
    \caption{Ratios of non-overlapped cache-related overhead over GPU kernel time within a single layer across different sequence lengths.}
    \label{tab:handling_overhead}
    \begin{tabular}{lcccccc}
        \toprule
             & 4k   & 8k   & 16k  & 32k  & 64k  & 128k \\
        \midrule
        H100 & 45\% & 51\% & 51\% & 59\% & 68\% & 63\% \\
        A40  & 31\% & 32\% & 33\% & 32\% & 41\% & 47\% \\
        \bottomrule
    \end{tabular}
\end{table}

\begin{figure}[t] 
    \centering
    \includegraphics[width=\columnwidth]{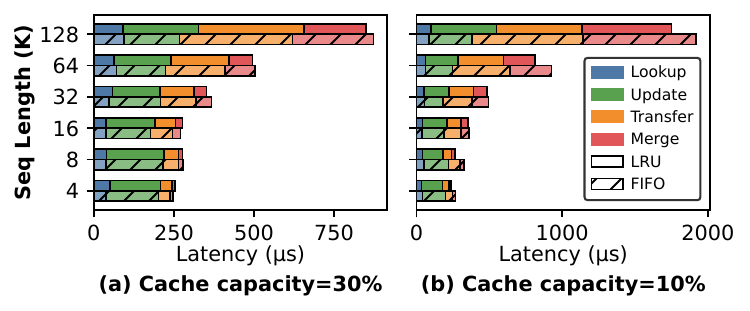}
    \caption{Cache-related timecost breakdown results of FIFO and LRU algorithms with different on-GPU cache sizes. H100 and RetroInfer are used.}
    \label{fig:background:handling_breakdown}
\end{figure}
Surprisingly, Figure~\ref{fig:background:n2n_latency} shows that both RetroInfer and PQCache underperform the two full attention variants across all evaluated settings, with their sparsity fixed at 10\% of total KV states. For example, RetroInfer runs 14–79\% and 15–143\% slower than FullAttn and FullAttn+CuGraph, respectively. PQCache performs significantly worse, lagging behind by 168–797\% and 176–1001\% relative to the two baselines. While enabling CUDA Graphs provides little benefit on the lower end A40 GPU, where kernel launching's overhead is moderate, the impact becomes substantial on the high end H100 GPU. For instance, FullAttn+CuGraph achieves a 1.01–1.75$\times$ speedup over its non-graph counterpart. Consequently, the performance gap between existing KVCache subsystems and FullAttn+CuGraph widens significantly on modern high performance GPUs. These inefficiencies lead us to conduct a deeper investigation, through which we identify two root causes and motivate a new KVCache subsystem design. Since PQCache consistently underperforms RetroInfer, we focus our subsequent analysis on RetroInfer.

\subsubsection{High on-GPU Cache Management Overhead} 
\label{sect:bg:mgtoverhead}

One key design philosophy behind modern KVCache subsystems is to improve the efficiency of the on-GPU cache so as to increase cache hit ratios and reduce the amount of data transferred over PCIe. However, these benefits come at the cost of substantial CPU-side cache management overhead. As illustrated in Figure~\ref{fig:background:cache_overview}, each decoding step triggers four major CPU assisted cache operations, namely, \texttt{lookup}, \texttt{transfer}, \texttt{update}, and \texttt{merge}, which collectively replenish the on-GPU cache. Among these, \texttt{update} operations can overlap with data transfer and GPU attention computation, whereas the remaining operations lie directly on the critical path of every inference request.

To quantify this overhead, we compare the non-overlapped cache management cost against GPU computation time (Table~\ref{tab:handling_overhead}). On A40, the ratio of cache-related time to GPU kernel execution ranges from 31\% to 47\% and increases with sequence length. This problem becomes far more pronounced on the H100, e.g., at a 128K sequence length, cache related overhead accounts for up to 63\% of the corresponding GPU compute time. This reflects the substantial periods during which the GPU remains idle, waiting for cache metadata and KV data to arrive from the CPU.

We further break down this overhead using RetroInfer’s LRU-based policy on H100, with the cache size set to 30\% of the full KVCache.  Figure~\ref{fig:background:handling_breakdown} (a) shows that \texttt{update}'s timeshares are 17-66\%, while \texttt{lookup} contributes only a small fraction (11-20\%), with both increasing modestly as sequence length grows. The high \texttt{update} cost stems from RetroInfer’s list based metadata structure. With a block size of 8 tokens, a 30\% cache size for 128K sequence requires each attention head to maintain a list of 4.8K entries. Promotions and evictions therefore require traversing long lists, resulting in substantial CPU overhead. In contrast, \texttt{transfer} and \texttt{merge} costs rise sharply with sequence length. Although LRU maintains a cache hit ratio above 92\%, longer contexts inevitably introduce more cache misses, causing \texttt{transfer} and \texttt{merge} to constitute a large fraction of total overhead at 128K.

A natural question is whether simpler cache algorithms can alleviate this overhead. Unfortunately, replacing LRU with FIFO offers no improvement. Figure~\ref{fig:background:handling_breakdown}~(a) depicts that FIFO resembles LRU across all settings and even performs slightly worse beyond 64K tokens. This is due to FIFO’s lower cache hit ratio (about 90\%, roughly 2\% lower than LRU), which increases miss induced \texttt{transfer} and \texttt{merge} overhead. Another potential optimization is increasing the block size to reduce metadata. However, enlarging RetroInfer’s block size from 8 to 128 produces no throughput gains. Although the overhead of \texttt{lookup} and \texttt{update} shrinks by under 10\%, the reduced cache hit ratio increases \texttt{transfer} and \texttt{merge} costs enough to negate any benefit.

We also evaluate a more memory-constrained scenario by reducing the on-GPU cache size from 30\% to 10\%. Figure~\ref{fig:background:handling_breakdown}~(b) shows that cache related overhead nearly doubles. The distribution across the four steps remains similar, but all steps except the negligible \texttt{lookup} grow more pronounced. LRU’s hit ratio drops to around 70\%, while FIFO falls below 60\%, leading to substantially more cache misses and correspondingly higher \texttt{transfer} and \texttt{merge} costs.

\noindent\textbf{Hint 1:} The on-GPU cache management for LLM KVCache subsystems requires \textbf{lightweight} designs that jointly consider cache granularity, hit ratio, miss penalty, and the unique access patterns of LLM decoding, rather than relying on generic cache algorithms.

\subsubsection{Difficulties of CPU-GPU Cooperation}
\label{sect:bg:sync}


LLM inference relies on close CPU–GPU cooperation, where the GPU performs nearly all computation, but the CPU remains responsible for launching every kernel and orchestrating data movement. As GPU compute capacity scales rapidly, this division of labor has become increasingly imbalanced. For example, while H100 delivers a 6.6$\times$ increase in FLOPS over A40, CPUs typically paired with them improve by only 1.29$\times$. This widening gap makes CPU overhead a dominant limiter in modern inference pipelines. The cost of launching kernels is particularly severe. In our H100 experiments, kernel launch overhead alone accounts for up to 60\% of end to end decoding latency.

To mitigate this gap, NVIDIA introduced CUDA Graphs~\cite{cuda-graph}, which capture a fixed topology of GPU operations including kernels, data movements, and their data dependencies into a directed acyclic graph that can be precompiled into a reusable executable. During inference, this enables the entire forward pass of an LLM to be launched through a single \texttt{cudaGraphLaunch} invocation, rather than dozens or hundreds of fine-grained kernel launches.  Figure~\ref{fig:background:n2n_latency} shows that enabling CUDA Graphs yields substantial performance improvement on high-end GPUs such as H100. 

However, CUDA Graphs impose strict constraints and require a static operation topology, stable tensor shapes and memory addresses, and the absence of any CPU side synchronization or dynamic control flow. As shown in Figure~\ref{fig:background:cache_overview}, each decoding step requires the CPU to inspect the cache state, determine which KV blocks must be fetched from host memory, wait for their arrival through asynchronous copies, and then inform the GPU the completion of data movement and which buffer slots to use. This procedure introduces many \texttt{cudaMemcpyAsync} calls, copy events, and CPU-side synchronizations into the GPU pipeline, creating fragmentation and preventing the entire iteration from being captured as a stable CUDA Graph.

\noindent\textbf{Hint 2:} High-end GPUs favor large, static, and self-contained GPU execution pipelines. An efficient KVCache subsystem must therefore avoid CPU-driven, fine-grained cache orchestration that produces fragmented operations and synchronization points incompatible with CUDA Graphs.

\section{Overview}
\label{sec:overview}


\begin{figure}[t]
    \centering
    \includegraphics[width=.8\columnwidth]{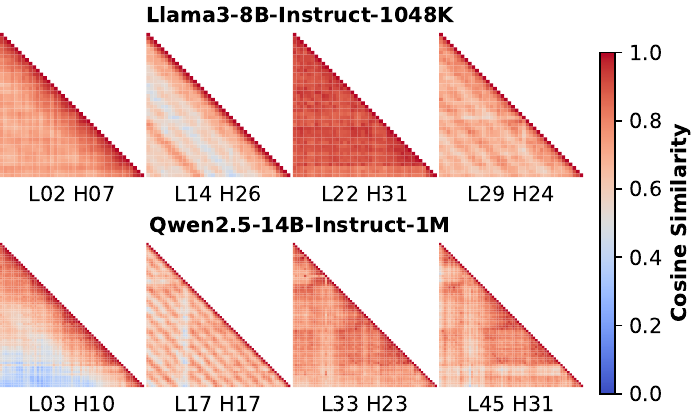}
    \caption{The cosine similarity distribution among query vectors. Evaluated on the decoding phase of a random sampled sequence. "L02 H07" means layer2 and head7. Adjacent queries (near the diagonal) exhibit high cosine similarity.}
    \label{fig:cache_design:cosine_distribution}
\end{figure}

\begin{figure}[t]
    \begin{subfigure}{.495\columnwidth}
        \centering
        \includegraphics[scale=0.465]{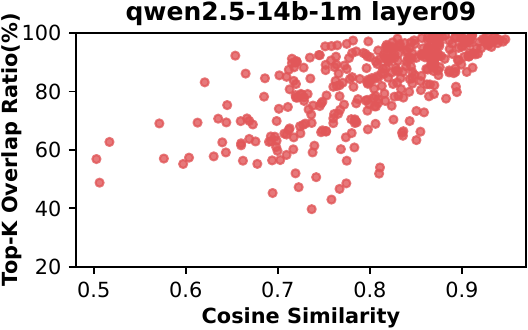}
    \end{subfigure}
    \begin{subfigure}{.495\columnwidth}
        \centering
        \includegraphics[scale=0.465]{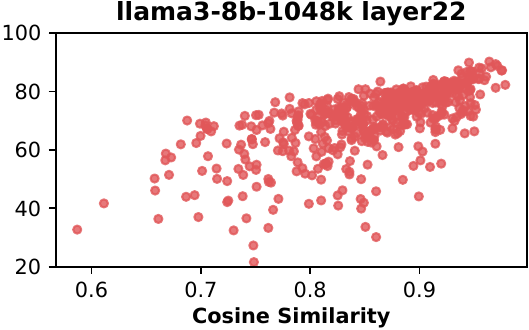}
    \end{subfigure}
    \caption{Correlation of the cosine similarity of adjacent query pairs and the overlap ratio of their \topk KV data. Evaluated on sequences randomly sampled from RULER-128K, measured at head granularity. \Topk ratio=10\%.}
    \label{fig:cache_design:topk_recall_by_cosine}
\end{figure}


\subsection{Key Observations}


Rather than relying on temporal locality of accessed \topk KV states as conventional cache algorithms do, our goal is to design a new cache algorithm that simplifies the decision of which \topk KV states are hot and how they should be replenished in the on-GPU cache. Our design is motivated by two LLM-specific observations, which together reveal an important property: \textit{adjacent queries from the same attention head tend to retrieve highly similar \topk KV states}.

First, note that the recent proposed advanced LLMs, such as Qwen~\cite{qwen2.5-1m}, Llama~\cite{touvron2023llama}, etc., adopt the rotary position embedding (RoPE) technique~\cite{su2024roformer}, under which the queries with adjacent positions will be rotated at similar angles. Consequently, we observe that the adjacent query vectors from the same attention head exhibit high cosine similarity, defined for any two query vectors $q_i$ and $q_j$ as 

\begin{equation*}
    \text{CosSim}(q_i,q_j) = \frac{\text{dot-product}(q_i,q_j)}{||q_i||*||q_j||},
\end{equation*}

A higher cosine similarity indicates closer alignment in vector direction. 
Using Qwen2.5 and Llama3, we compute cosine similarities for all query pairs within the same inference request and the same head. 
As shown in Figure~\ref{fig:cache_design:cosine_distribution}, query pairs located near the diagonal,  corresponding to queries generated in adjacent decoding steps, consistently show high cosine similarity.

Second, this similarity is strongly reflected in the KV states they retrieve. In Figure~\ref{fig:cache_design:topk_recall_by_cosine}, we correlate cosine similarity of adjacent queries with the overlap ratio of their retrieved \topk KV states. Both Qwen2.5 and Llama3 exhibit a clear positive correlation, where higher cosine similarity leads to greater overlap in the retrieved KV states. Especially, when cosine similarity exceeds 0.9, the two queries share the vast majority of the \topk KV states. We observe the same phenomenon in other widely used models, including DeepSeek~\cite{deepseekai2024deepseekv2strongeconomicalefficient}, GLM~\cite{glm2024chatglmfamilylargelanguage}, and GPT~\cite{openai2025gptoss120bgptoss20bmodel}, and provide additional evidence in the Appendix.A.

In fact, this phenomenon is in line with the \topk selection mechanism.  Note that given a target query $q$,  \topk attention needs to compare the \textit{qkscores}  between the query $q$ and all the keys. 
For an arbitrary $k_i$,  the \textit{qkscore}  is measured by 
\begin{equation}
    \text{dot-product}(q, k_i) = q^\top k = ||q||*||k_i||cos(q, k_i).
\end{equation}
Though the qkscore is related to both the vector direction and the vector length of $q$, the relative ordering of qkscores between q and all keys $\{k_i\}$ is independent of $||q||$ since $||q||$ is shared. As a result, two queries with similar vector directions correspond to similar ordering of qkscores and the retrieved KV states should overlap to a large extent. 

\begin{figure}[!t]
\centering
    \includegraphics[width=\columnwidth]{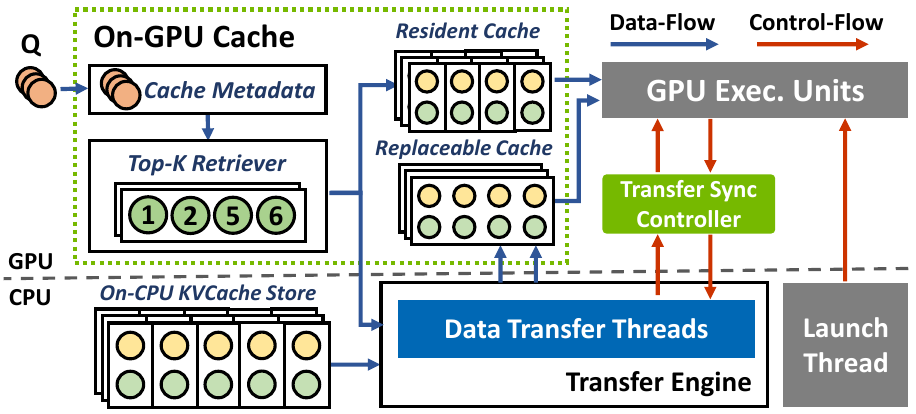}
    \caption{System architecture of \sysname.}
\label{fig:design:system_overview}
\end{figure}

\subsection{Design Rationale}
\label{sect:design:rationale}




The above strong intra-head similarity of adjacent queries motivates us to design a head-wise KVCache mechanism, which is called query similarity-based approximate cache (or short, \algoname). It reuses KVCache entries across decoding steps whenever query directions remain similar, dramatically reducing cache management overhead.

\algoname eliminates the list based metadata structures used in prior work. Instead, each head retains only the query vector from the previous decoding step as its metadata, resulting in a single lightweight entry per head. Cache lookup becomes a cosine similarity check, implemented as a single efficient GPU kernel. Cache update simply replaces the previous step’s stored query with the current decoding-step one upon cache miss, implemented as another GPU kernel. Because each head either keeps its entire \topk KV data in GPU memory or relies entirely on CPU memory, the system no longer requires merge operations. In the end, our new design reduces cache management to two simple GPU-side operations with no CPU involvement, enabling a much more efficient and predictable cache pipeline. Furtheremore, this GPU-centric design makes \algoname more friendly with CUDA Graphs' execution mode. The detailed design of \algoname refers to Section~\ref{sec:cache-algo}.

To realize the benefits of the \algoname, we must address two key technical challenges.

\noindent \textbf{Challenge 1: Balancing accuracy and performance under similarity based reuse.} 
\algoname relies on a similarity threshold to decide when a query can safely reuse the cached head level KV data from the previous step. A low threshold allows aggressive reuse but risks degrading inference accuracy. A high threshold increases accuracy but leads to frequent cache misses, reducing performance. Determining an appropriate threshold that achieves both accuracy and efficiency is therefore nontrivial. Additionally, not all queries exhibit strong similarity even when their retrieved KV data remain important. These corner cases must be handled carefully to prevent accuracy loss without compromising cache efficiency.

\noindent\textbf{Challenge 2: Managing CPU-GPU coordination under increased data movement.} 
Although \algoname eliminates CPU-side metadata management and shifts most cache logic to the GPU, it does not remove all CPU-GPU interactions. Cache misses still require fetching \topk KV data from CPU memory, and the GPU must be correctly notified when these transfers complete. Moreover, by trading cache hit ratio for significantly lower cache management overhead, \algoname may cause more KV data to be transferred from CPU to GPU. These factors collectively make it difficult to maintain a clean and static execution pipeline suitable for CUDA Graphs.

\subsection{Overall System Architecture}
\label{sect:sysarchi}

To address the above challenges, building on our \algoname algorithm, we introduce our new KVCache subsystem \sysname that enables more efficient LLM inference, illustrated in Figure~\ref{fig:design:system_overview}. The system preserves the familiar two-tier architecture, where the CPU-side KVCache store continues to manage KV data at token or block granularity as determined by the specific \topk algorithm. The kernel launch thread and GPU execution units remain the same as the existing systems such as RetroInfer. Beyond this, our system-wise innovations lie in the following added components that enable lightweight on GPU cache management:


\noindent \textbf{On-GPU Cache} maintains two cache buffers on GPU. The first is the \textit{Replaceable Cache}, which stores the \topk KV data at head granularity and is sized to hold the \topk portion of the full KVCache. The second is the \textit{Resident Cache}, which keeps KV blocks that exhibit low similarity yet remain important (Section~\ref{sec:outlier_handling}), using the same granularity as the CPU-side KVCache store. In addition, the module employs a \textit{\Topk Retriever}, allowing \sysname to perform efficient KV selection with existing \topk attention algorithms. 

\noindent \textbf{Data Synchronization Controller.} To be compatible with CUDA Graph, we move synchronization logic from CPU to GPU. The controller performs GPU side synchronization and stalls only GPU execution when needed, while a dedicated CPU thread continuously issues kernel launches. This design avoids fine grained CPU coordination and preserves the static, bulk execution pattern required by CUDA Graph.

\noindent \textbf{Data Transfer Engine.} This component executes the CPU-to-GPU KV data transfer. It employs several dedicated CPU Data Transfer Threads to move required \topk KV data to the Cache Buffer in GPU HBM via zero-copy operation.
\section{Cache Algorithm Design}
\label{sec:cache-algo}
\noindent Here, we present \algoname algorithm that exploits strong intra-head query similarity to eliminate costly cache operations.

\begin{algorithm}[t!]
    \caption{\algoname}
    \label{alg:qsac}
    \begin{algorithmic}[1]
        \Require The threshold $\tau_i$ for each attention head $i$.
        
        \State Initial: at step $0$, each head $i$ caches $q^i_c=q^i_0, K^i_c=[k^i_0], V^i_c=[v^i_0]$.
        
        \For{$1\leq t \leq T$}
            \State Given the states $q^i_t, k^i_t, v^i_t$ of time step $t$.
            \State Compute query similarity $\text{CosSim}(q^i_t, q^i_c)$, for each head $i$. \Comment{cache lookup}
            
            \If {$\text{CosSim}(q^i_t, q^i_c)>\tau_i$} \Comment{cache hit} 
                \State Loading KV from the GPU cache.
            
            \Else \Comment{cache miss} 
                \State Compute $\topk$ indices by
                \State \quad $\text{index}_i = \text{TopkRetrieval}(q^i_t, K^i, V^i)$
                \State Obtain the $\topk$ KV data by
                \State \quad $\hat{K}^i=\textbf{Gather}(K^i, \text{index}_i)$
                \State \quad $\hat{V}^i =\textbf{Gather}(V^i, \text{index}_i)$
                \State Transfer KV data by $K^i_c=\hat{K}^i, V^i_c=\hat{V}^i$.
                \State Update query cache by $q^i_c=q^i_t$.  \Comment{cache update}
            \EndIf
            
            \State Compute the attention output by
            \State \quad $\text{output} =\text{Attn}(q^i_t, K^i_c, V^i_c)$
        \EndFor
    \end{algorithmic}
\end{algorithm}

\subsection{Query Similarity-based Approximate Cache}
\label{Sec:cache_design:obversation}
The main idea is to  reuse the KVCache for the adjacent steps
when the queries exhibit high cosine similarity. To this end, we need to manage the KVCache in a head-wise manner.
Algorithm~\ref{alg:qsac} shows the workflow of our design. 

Compared with the existing KVCache subsystems,  except the KV data cached on GPU, we further cache a query vector $q^i_c$ as metadata for each head $i$. 
In \algoname, the lookup becomes a simple cosine similarity computing between the current query $q^i_t$ and the cached query $q^i_c$ (Line 4). 
If the similarity exceeds the threshold ${\tau_i}$, it indicates a cache hit (Line 5). 
In this case, $q^i_t$ is considered a close neighbor of $q^i_c$, and the \topk KV data can be directly reused and loaded from Replaceable Cache (Line 6). 
Otherwise, $q^i_t$ is far different from $q^i_c$, which means cache miss (Line 7).  
In this situation, we retrieve the \topk KV states for $q^i_t$ and update both the cache buffer for KV (Line 10-13) and query metadata (Line 15).  
Note that, \algoname does not include the buffer merge step, because the required KV states per head are obtained totally from the GPU if cache hit, and from the CPU if cache miss, rather than using a CPU-GPU KVCache mixture as in the existing systems. 

It is worth noting that we update the query only on cache misses. 
This ensures alignment between the query and the cached KV data. 
Updating the query at every step without refreshing the KV data would gradually introduce mismatches between the cached KV and the query representations, ultimately degrading inference accuracy.


\subsection{GQA-Aware Similarity Aggregation}
\label{sec:gqa_aggregation}
To align with the widely-used GQA, data loading and caching must operate at KV-head granularity. However, cosine similarity computation occurs per query head. As a result, an effective intra-GQA similarity aggregation method is required to bridge the gap between KV heads and query heads.

In order to apply our method to GQA scenarios, we need first define the importance of each query head. Though ~\cite{xiao2025duoattention} only gave the method to compute the importance score of KV heads, we show that such a method can also be utilized to compute the importance score of query heads, with a slight modification. The detailed process is given in Appendix.B. Now we can give our intra-GQA aggregation method based on the query head importance. To ensure accuracy, we adopt a conservative aggregation manner that follows the two rules:
\begin{itemize}
\item The query head with larger importance score should  account for a higher proportion during the aggregation.
\item The aggregation result should be mainly determined by the query heads with low cosine similarities, especially when they have large importance scores.
\end{itemize}
Considering a query group including $m$ query heads, denoted by $q_1, q_2, ..., q_m$, whose importance scores are $s_1, s_2, ..., s_m$. Assume that at one step, the cosine similarities are $\text{sim}_1, \text{sim}_2, ...,$ $ \text{sim}_m$, respectively, and then
the aggregated similarity $s$ is expressed by
\begin{equation}
    s = \frac{s_1+s_2+...+s_m}{\frac{s_1}{\text{sim}_1}+\frac{s_2}{\text{sim}_2}+...+\frac{s_m}{\text{sim}_m}}.
\end{equation}

Such an aggregation satisfies the above two rules, and in addition, when all the query heads have the same cosine similarities, the  aggregation result maintains the same value, which is reasonable.

\subsection{Similarity Threshold Tuning}
\label{sec:threshold_tuning}

The thresholds $\tau_i$ are important hyper-parameters that affect both accuracy and performance of the proposed approximate caching method. A lower value of $\tau_i$  implies high-frequency reuse of the cached data and low-frequency cache update, contributing to a high inference efficiency.  However, in this situation, the query may miss actually important KV states, which leads to accuracy degradation. On the contrary, a higher value of  $\tau_i$ may achieve a high accuracy but a low efficiency. Therefore, how to configure the reuse thresholds to balance performance and accuracy emerges as the pivotal challenge.

Duo-Attention~\cite{xiao2025duoattention} observed the unequal contributions of KV heads to model accuracy. It also proposed a methodology to profile and quantify this contribution, which produces \textit{head importance scores} in the range from $0$ to $1$, and  heads with larger importance scores exhibit higher sensitivity to KV data.  Inspired by this, we aim to leverage \textit{head importance} to address the above-mentioned challenges.

\noindent \textbf{Adaptive threshold configuration.} We assign a larger threshold to the attention head with larger importance score to protect its effective information. First, we select an upper bound threshold $\eta$ for all heads. Then, for an attention head $i$ with importance score $s_i$, its reuse threshold $\tau_i$ is computed by
\begin{align}
    & \theta^* = \text{arccos}(\eta) \label{Equation:cache_design:thresholds1} \\
    & \theta_{i} = \lambda_i\theta^* + (1-\lambda_i)\pi  \label{Equation:cache_design:thresholds2} \\
    & \tau_i = \text{cos}(\theta_{i}) \label{Equation:cache_design:thresholds3}
\end{align}
\noindent
Here, Equation (\ref{Equation:cache_design:thresholds1}) converts similarity threshold upper bound $\eta$ into an angle threshold upper bound $\theta$. In Equation (\ref{Equation:cache_design:thresholds2}), the coefficient $\lambda_i =  s_i^p$, which determines the actual angle threshold $\theta_i$ for head $i$ according to its importance.  Equation (\ref{Equation:cache_design:thresholds3}) yields the final similarity threshold. In practice, one can set $p$ to 2 or 3 in which case most attention heads have a high reuse rate while the truly important heads can be timely updated, ensuring both effectiveness and efficiency.

\subsection{Outlier Handling}
\label{sec:outlier_handling}

The proposed cache policy may encounter some hard-to-reuse heads that involve  low hit rates. For  such heads, the KVCache in HBM needs to be frequently updated, which will affect the overall efficiency.  
To tackle this issue, we propose \textit{resident caching}, which retains the full KVCache of the hard-to-reuse heads in GPU HBM. Before detailing this technique, we need to first quantify the \textit{reuse difficulty} of the KV heads:


\noindent \textbf{Quantifying reuse difficulty.} 
We define the reuse difficulty of a KV head, denoted by $D_i$, as $D_i = \tau_i - (\hat{s}_i - \epsilon)$,
where $\tau_i$ is the similarity threshold of this KV head and $\hat{s}_i$ is its offline profiled average cosine similarity, which is evaluated on adjacent query pairs generated by tens of random sequences. $\epsilon > 0$ is an error tolerance margin.

\noindent \textbf{Resident caching.} We select heads with highest $D_i$ which should be most hard-to-reuse and retain their full KVCache in GPU HBM. This is feasible because the memory size of our cache data buffer equals the \topk KV data size (e.g., only 10\% of the total KVCache), leaving substantial free capacity in GPU HBM. In practice, we will set a moderate number of heads with resident caching, according to the specific remaining HBM capacity.


\section{CPU-GPU Cooperation}
\label{sec:cpu-gpu-cooperation}

\subsection{Cache-Miss Acceleration}

In last section, we introduce the cache algorithm QSAC  embedded in our system LiteCache, which greatly reduces the cache related overhead. In this section we equip LiteCache with the recently proposed systematic optimization speculative sparse prefetching proposed in  InfiniGen~\cite{lee2024infinigen} to further improve the efficiency.

Speculative sparse prefetching provides a next-layer query predict trough which the next-layer PCIe transfer and current-layer inference computation can be overlapped to some extent.  This technology can be well combined with our QSAC cache mechanism. Note that InfiniGen does not involve on gpu cache and it requires to transfer the \topk KVCache of all heads. Such a large amount of transfer is far beyond the inference computation, so that the overlapped transfer time is limited. While in the QSAC cache mechanism, we only prefetch the hard-to-reuse heads to accelerate our cache miss cases. In addition, some hard-to-reuse heads have been addressed by resident caching in Section~\ref{sec:outlier_handling}, so only a few unaddressed heads requires prefetching. The prefetching data is greatly reduced and the transfer time can be well overlapped.

Speculative  sparse prefetching and  QSAC utilize the query similarity in adjacent layers and adjacent steps, respectively. The two characteristic are orthogonal and can be well combined. Equipped with the two techniques, LiteCache maximizes the capabilities of GPU chips.
\subsection{GPU-Centric Synchronization}
\begin{figure}[t!]
\centering
    \includegraphics[width=\columnwidth]{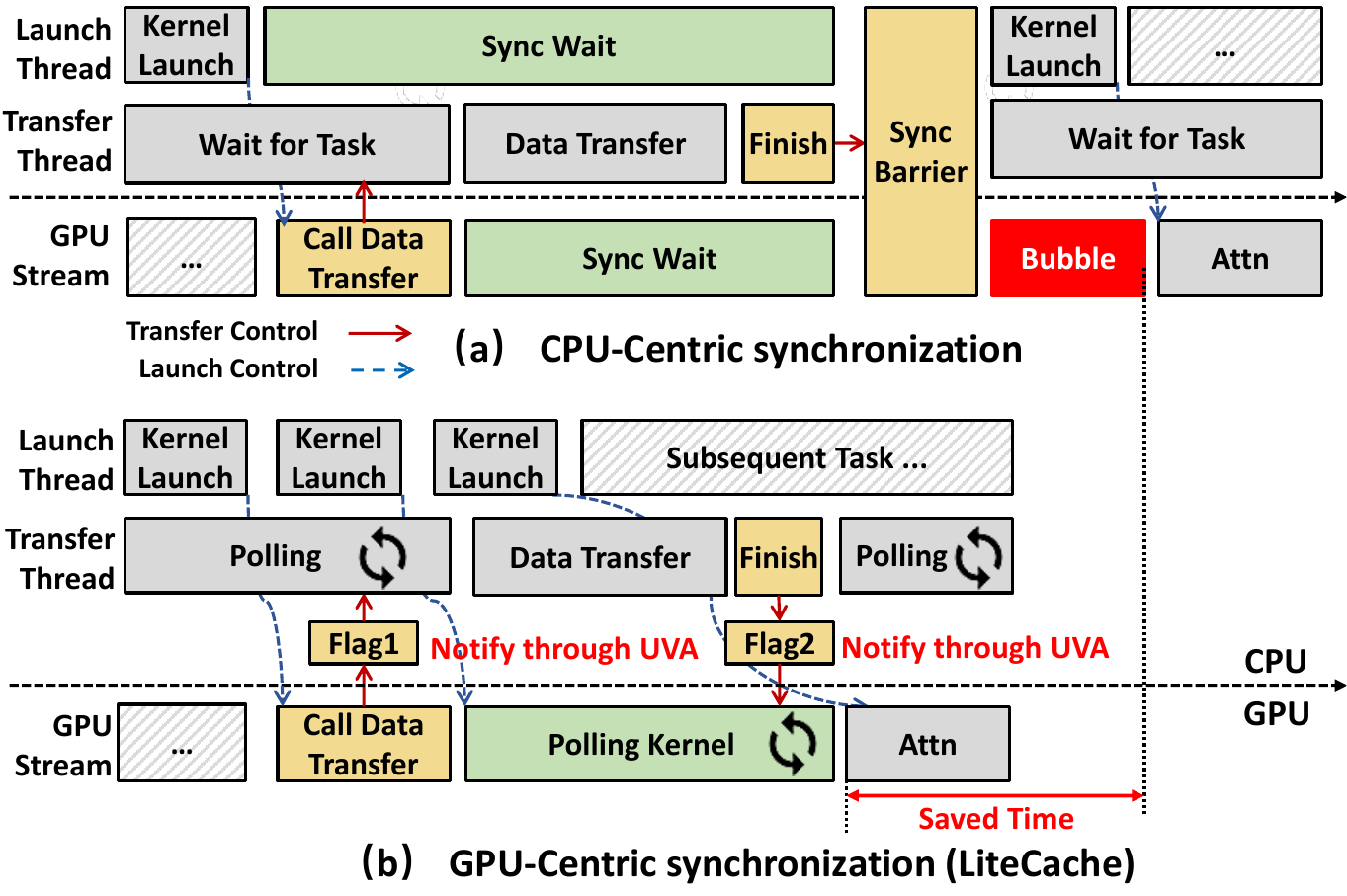}
    \caption{CPU-centric and GPU-centric synchronization.}
\label{fig:design:Sync&Async}
\end{figure}

To correctly maintain CPU–GPU coordination, existing systems rely on a CPU-centric synchronization strategy. 
As illustrated in Figure~\ref{fig:design:Sync&Async}~(a), they insert a barrier whenever synchronization is required, which stalls both the CPU and GPU. 
The CPU is prevented from launching subsequent kernels, creating GPU bubbles. 
Moreover, because the barrier is executed on the CPU and only becomes visible to the GPU at runtime, it cannot be statically captured by CUDA Graph.

To address this problem, we design a GPU-centric synchronization mechanism. As shown in Figure~\ref{fig:design:Sync&Async}~(b),
which blocks the GPU execution instead of the CPU, thus ensures that CPU threads can continuously launch GPU kernels.
We decouple kernel launching from data transmission and synchronization tasks by assigning them to a separate CPU thread. 
The data synchronization is confirmed via flag variables in CPU-GPU shared memory. 
Specifically, when a transfer is requested, the GPU writes data indices to host memory through Unified Virtual Addressing (UVA)~\cite{nvidia_uva} and sets the corresponding \texttt{Flag1} in shared memory to trigger the transfer engine. 
Upon completion of the transfer, the transfer engine updates these \texttt{Flag2} to notify the GPU. 
The GPU polls this flag via a \texttt{Polling Kernel} and proceeds subsequent kernels only after confirming that the data transfer has finished.

This design prevents CPU-side kernel launches from being stalled by transfer synchronization, and the polling kernel runs entirely on the GPU, allowing it to be statically captured by CUDA Graph. 
As a result, \sysname can fully leverages the advantage of CUDA Graph for inference.
\section{Put It All Together}
\label{sec:subsystem}
\noindent We incorporate the above coherent design into a KVCache subsystem \sysname, to enable efficient LLM inference. This section presents its workflow and implementation.

\subsection{\sysname System Workflow}
\label{sec:subsystem:workflow}
The workflow of \sysname can be divided into three phases.

\noindent \textbf{Initialization phase.} During initialization, the KVCache heads are partitioned into HBM-resident and offloaded groups following the strategy described in Section~\ref{sec:cache-algo}. 
This partitioning is guided by offline-profiled statistics, including head importance, average cosine similarity, and the number of prefetchable heads.

\noindent \textbf{Prefill phase.} \sysname introduces two operations in the prefill phase. 
First, following the \topk attention algorithm it adopts, \sysname encodes the KCache into \topk retrieval metadata as presented in Section~\ref{Sec:background:kvoffloading}, which incurs minimal overhead compared to other computations in this phase~\cite{gong2025hata}. 
Second, it offloads partial or full KV heads to host memory according to the initialization partitioning results. This offloading is asynchronously executed alongside GPU computation tasks, with its latency fully hidden by computation.

\begin{figure}[t!]
\centering
    \includegraphics[width=\columnwidth]{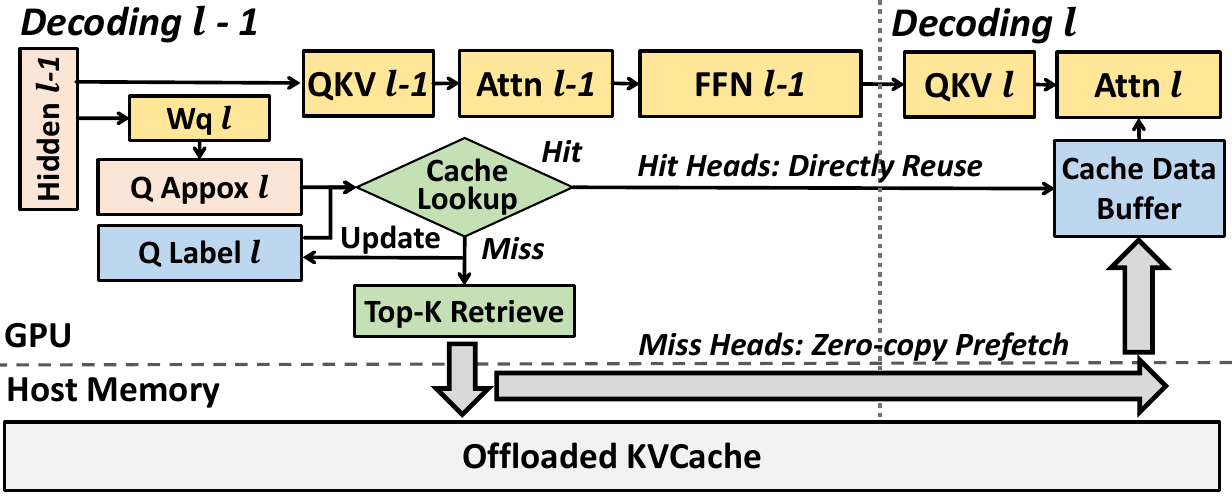}
    \caption{Inference workflow of \sysname during the decoding phase. For clarity, only processes relevant to cache accessing and host data prefetching for layer $l$ are depicted.}
\label{fig:design:decoding_workflow}
\end{figure}


\noindent \textbf{Decoding phase.} 
During decoding, \sysname dynamically retrieves and loads the \topk KV data for each step. 
Figure~\ref{fig:design:decoding_workflow} illustrates this workflow. For layer $l$, \sysname begins to retrieve and load its \topk KV data at the beginning of layer $l-1$. 
It computes an approximate query vector using the method in InfiniGen~\cite{lee2024infinigen}, and uses it for cache lookup to identify hit heads.
For missed heads, \sysname retrieves new \topk KV data and launches the prefetch operation. 
Then, it continues to execute the remaining computation of layer $l-1$ (attention, FFN). 
Note that the prefetching mechanism splits the cache update into two phases, namely \textit{query labels update} and \textit{KV data update}. 
Query labels for missed heads are immediately updated upon the finish of cache lookup, while their KV is gradually updated during prefetching. 
After the qkv projection of layer $l$, \sysname retrieves \topk KV data from Append-only buffer using layer $l$'s actual query vector. 
Finally, after prefetching completes, layer $l$'s attention computation is performed using data from the Cache Data Buffer, which consists of Append-only Cache and Replaceable Cache, as shown in Figure~\ref{fig:design:system_overview}.

\subsection{Implementation Details}
\sysname is implemented based on FlashInfer~\cite{ye2025flashinfer} and Transformers~\cite{wolf-etal-2020-transformers}, with 3,417 lines of CUDA/C++ code and 1,555 lines of Python code. 
Beyond the aforementioned efficient system design and optimizations, \sysname has also integrated the following additional optimizations to further improve the inference performance:

\para{Zero-copy transfer engine.}
\sysname employs a zero-copy engine for CPU–GPU KV transfer that eliminates redundant CPU-side gathering, maximizes PCIe bandwidth, and reduces transfer latency. 
The engine is built on GDRCopy~\cite{gdrcopy}, which exposes GPU memory through the PCIe BAR~\cite{PCIe-Bar} region and enables direct CPU access via load–store instructions. 
We use a thread pool to transfer KV data at the granularity of a single K/V vector, whose 128 FP16 elements align with the 64-byte PCIe transfer unit. 
To further boost zero-copy transfers, the engine exploits AVX SIMD instructions~\cite{avx} and CPU cache prefetching~\cite{cpu-cache-prefetch} to improve throughput and mitigate random-access stalls. 
Overall, this design substantially accelerates KV data movement across the CPU–GPU boundary.

\para{Kernel fusion for cache lookup and query label update.} 
As described in Section~\ref{sec:subsystem:workflow}, in \sysname's workflow, query label update immediately follows the cache lookup. 
Both operations require to access the incoming query vector and the historical query labels. 
To avoid redundant HBM accesses and reduce CPU-side kernel-launch overhead, we fuse these two operations into a single GPU kernel.

\para{Hybrid attention kernel.} 
\sysname's \topk attention computation requires accessing data from both \texttt{Append-only Cache} and \texttt{Replaceable Cache}. 
To accommodate this hybrid data layout, we implement an attention kernel based on FlashAttention2~\cite{dao2024flashattention}, which directly reads data from both buffers, without the need for a separate data gathering operation. 
This design effectively reduces HBM traffic and lowers memory-access overhead.

\begin{figure*}[h]
    \centering
    \includegraphics[width=\textwidth]{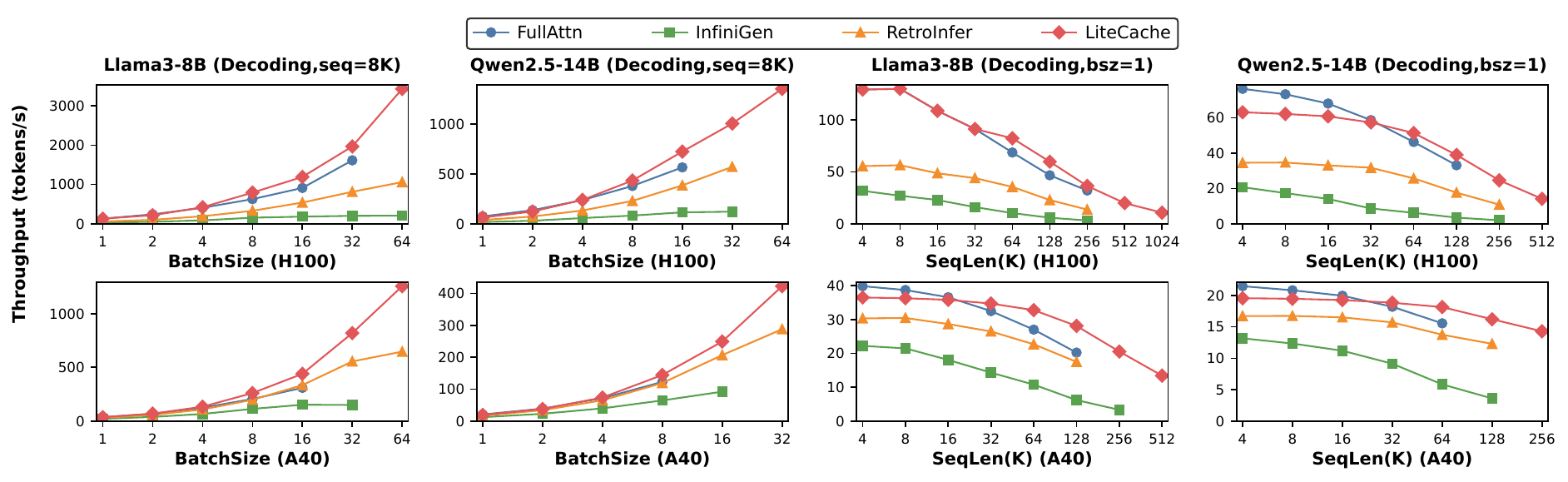}
    \label{fig:throughput:a40}
    \caption{End-to-end decoding throughput across varying batch sizes and sequence lengths.}
    \label{fig:throughput}
\end{figure*}
\section{Evaluation}
\label{sec:eval}

\subsection{Experimental Setup}
\label{sec:eval:setup}




\noindent \textbf{Platform and models.} 
We conduct evaluations on two machines.
One machine is equipped with an AMD EPYC 9654 96-core CPU and an NVIDIA A40 GPU (48 GB) connected via PCIe 4.0 $\times$16, while the other is with an Intel(R) Xeon(R) Platinum 8457C 90-core CPU and an NVIDIA H100 GPU (80 GB) connected via PCIe 5.0 $\times$16.
The main results of two open-source LLMs are shown: Llama3-8B-Instruct-1048K~\cite{gradientlongcontextllama3} (32 layers, 32 Q heads and 8 KV heads per layer) and Qwen2.5-14B-Instruct-1M~\cite{qwen2.5-1m} (48 layers, 40 Q heads and 8 KV heads per layer).

\noindent \textbf{Benchmarks.} We evaluate \sysname on two widely used benchmarks. The first is \textit{RULER}~\cite{hsieh2024ruler}, which consists of 13 subtasks covering retrieval, multi-hop tracking, aggregation, and question answering, with configurable context lengths. The second is \textit{LongBench}~\cite{bai-etal-2024-longbench}, a comprehensive benchmark that includes diverse real-world tasks such as summarization, question answering, and code comprehension, spanning a wide range of input text lengths.

\noindent \textbf{Baselines and configurations.}
We compare \sysname with SOTA KVCache offloading systems, RetroInfer~\cite{chen2025retroinfer} and InfiniGen~\cite{lee2024infinigen}. RetroInfer adopts an LRU GPU cache to reduce data transfer, while InfiniGen hides transfer overhead via prefetching. 
We also include the full-attention implementation without offloading to highlight \sysname’s advantages. 
We further consider PQCache~\cite{zhang2025pqcache}. However, it exhibits higher latency in our experiments, mainly due to heavy cache-management overhead and a low cache-hit ratio (Section~\ref{Sec:background}).
Given its stronger performance, RetroInfer serves as the primary on-GPU caching baseline in our evaluation.
For all offloading baselines, we fix sparsity at 10\% of the total KV states, following InfiniGen, a common setting~\cite{tang2024quest, singhania2024loki,gao2025seerattentionlearningintrinsicsparse}.
Unlike PQCache and RetroInfer, LiteCache can work with various top-k attention algorithms~\cite{gong2025hata, tang2024quest, singhania2024loki}. We use HATA [15] (hash bits = 256) for evaluation due to its strong accuracy, but other top-k attention algorithms lead to similar conclusions under the same sparisity, as they only affect the lightweight top-k retrieval step.
For \sysname-specific hyperparameters, we fix $\eta = 0.8$, $p = 3$ in $\lambda_i = s_i^p$, and set $\epsilon$ to $0.1$ (Llama3-8B) and $0.05$ (Qwen2.5-14B). These values are selected from a small validation sweep and used in all experiments. Details are provided in Appendix.C.


\para{Sink and recent tokens.} Suggested by Duo-Attention~\cite{xiao2025duoattention}, even for KV heads with zero importance, sink and recent tokens remains necessary for preserving model accuracy. 
Following this, we persistently keep 4 sink tokens and 64 recent tokens in cache, which is a configuration commonly adopted in such scenarios~\cite{chen2025retroinfer,xiao2024streamingllm}.

\para{CUDA Graph support.}
Considering the performance benefits of CUDA Graph, we enable it for \sysname and full-attention. 
Other baselines, such as RetroInfer and InfiniGen, do not support CUDA Graph by design due to their dynamic and CPU-centric  execution patterns, and thus are evaluated without CUDA Graph support.

\subsection{Decoding Throughput Comparison}

We first present the decoding throughput of different methods on both NVIDIA H100 and A40 GPUs in Figure~\ref{fig:throughput}. Some results of baselines are missing due to GPU out-of-memory (OOM) errors 
of their open-source implementations.

Overall, \sysname consistently outperforms other offloading-based baselines on both H100 and A40. 
On H100, it delivers a speedup of 2.08-3.24$\times$ on Llama3 and 1.67-2.27$\times$ on Qwen2.5 compared to the leading baseline. 
On A40, the speedup can be up to 1.61$\times$ and 1.47$\times$ on Llama3 and Qwen2.5, respectively.
The throughput of \sysname scales significantly with increasing batch size, consistently outperforming other methods on both models. For short sequences, FullAttn is the top performer because attention is not the primary bottleneck, and other systems incur \topk retrieval overhead. 
However, as the sequence length grows, \sysname maintains higher throughput. 
InfiniGen shows the lowest throughput due to significant data-transfer overheads that cannot be fully hidden, while RetroInfer is increasingly slowed by heavy cache-related overheads.

\begin{figure}[t!]
    \centering
    \includegraphics[width=\columnwidth]{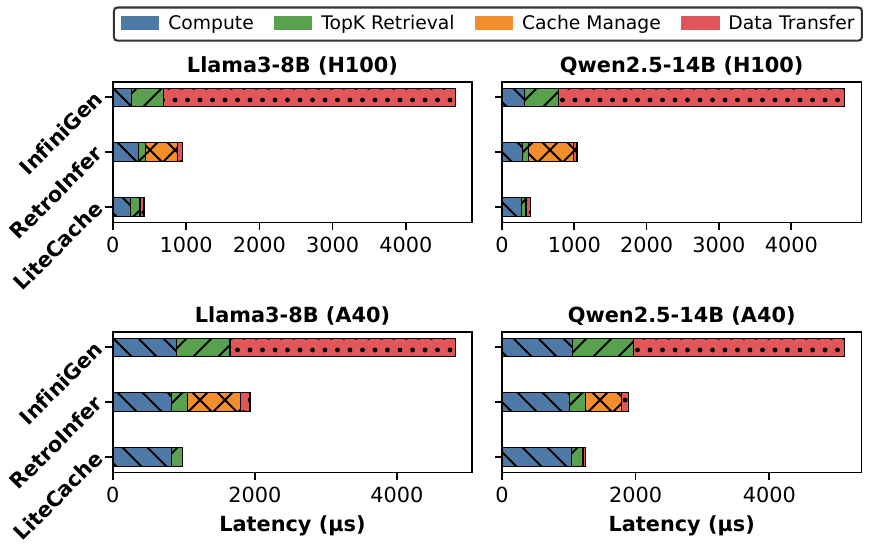}
    \caption{Per-layer decoding latency breakdown, evaluated with batch size=1 and sequence length=128K.}
    \label{fig:eval:breakdown}
\end{figure}

\subsection{Breakdown Analysis}

We further breakdown the per-layer decoding latency of \sysname, InfiniGen, and RetroInfer to better understand the sources of throughput differences.
Note that a key efficiency advantage of \sysname is its compatibility with CUDA Graphs, which substantially reduces kernel launch overhead in the decoding step on high-end GPUs. 
However, RetroInfer and InfiniGen cannot be integrated with CUDA Graphs due to their dynamic, CPU-centric execution patterns.
To keep the breakdown focused on the intrinsic differences in KV-offloading design, we do not include kernel launch overhead in this section. We will present an ablation study of the advantage of CUDA Graph on \sysname in section~\ref{sec:eval:new_ablation}.


As shown in Figure~\ref{fig:eval:breakdown}, \sysname achieves a 2.0–4.9$\times$ speedup on A40 and 2.2–16.1$\times$ on H100 compared to RetroInfer and InfiniGen, demonstrating its efficiency. 
Cache management overhead in \sysname is minimal due to its lightweight caching strategy, whereas RetroInfer spends 28.5-60.2\% of decoding time on cache management, and InfiniGen has no cache management by design. 
The data-transfer overhead is low for both RetroInfer and \sysname. RetroInfer benefits from a high cache-hit ratio, while \sysname uses speculative sparse prefetching to overlap the transfer of small miss-induced KV states with computation.
InfiniGen also uses speculative prefetching, but the large CPU-to-GPU data volume hinders full overlap.


\begin{table}[t!]
\centering
\setlength{\tabcolsep}{2.1pt}
\renewcommand{\arraystretch}{1.1}
\footnotesize
\caption{GPU cache memory usage (GB) vs. total tokens}
\begin{tabular}{l|ccccc|ccccc}
\toprule
\multirow{2}{*}{\textbf{Methods}} 
& \multicolumn{5}{c|}{\textbf{Llama3-8B, \#tokens}} 
& \multicolumn{5}{c}{\textbf{Qwen2.5-14B, \#tokens}} \\ \cline{2-11}
& 32K & 64K & 128K & 256K & 512K
& 16K & 32K & 64K & 128K & 256K \\
\midrule
RetroInfer & 1.20 & 2.40 & 4.80 & 9.56 & 19.12   & 0.90 & 1.79 & 3.59 & 7.20 & 14.34 \\
\sysname  &0.53 & 1.09 & 2.62 & 8.19 & 14.09 & 0.37 & 0.74 & 1.49 & 3.44 & 8.03 \\
\bottomrule
\end{tabular}
\label{tab:cache_mem_usage}
\end{table}

\begin{table}[t!]
\centering
\setlength{\tabcolsep}{2.2pt} 
\renewcommand{\arraystretch}{1.1} 
\footnotesize 
\caption{Cache hit ratio (\%)}
\begin{tabular}{l|ccccc|ccccc}
\toprule
\multirow{2}{*}{\textbf{Methods}} 
& \multicolumn{5}{c|}{\textbf{Llama3-8B, SeqLen=8K}} 
& \multicolumn{5}{c}{\textbf{Qwen2.5-14B, SeqLen=8K}} \\ \cline{2-11}
& 4 & 8 & 16 & 32 & 64
& 2 & 4 & 8 & 16 & 32 \\
\midrule
RetroInfer     & 90.4& 90.1& 90.1& 90.0& 90.1&  92.7&  91.3&  91.4&  91.3&  91.4\\
\sysname  & 79.5& 74.7& 76.7& 85.0& 80.0& 80.6& 80.7& 80.7& 81.5& 80.8\\
\midrule
\multirow{2}{*}{\textbf{Methods}} 
& \multicolumn{5}{c|}{\textbf{Llama3-8B, BSZ=1}} 
& \multicolumn{5}{c}{\textbf{Qwen2.5-14B, BSZ=1}} \\ \cline{2-11}
& 32K & 64K & 128K & 256K & 512K
& 16K & 32K & 64K & 128K & 256K \\
\midrule
RetroInfer     & 91.2& 91.4& 91.4& - & - & 93.1& 93.1& 93.4& 93.1& -\\
\sysname  & 83.5& 82.6& 84.0& 90.1& 88.6& 75.8& 73.9& 65.2& 72.5& 68.3\\
\bottomrule
\end{tabular}
\label{tab:hit_ratio}
\end{table}

\subsection{On-GPU Cache Statistics}

To better understand the effect of on-GPU caching, we compare \sysname and RetroInfer on the A40 platform.
Both systems maintain an on-GPU cache to reduce PCIe data transfers.
Table~\ref{tab:cache_mem_usage} shows GPU cache memory usage, while Table~\ref{tab:hit_ratio} reports the cache hit ratios corresponding to the A40-based experiments in Figure~\ref{fig:throughput}.
\sysname uses substantially less GPU memory for cache, consuming on average only 47.6\% of the memory required by RetroInfer.
Although RetroInfer attains a higher cache hit ratio, this causes the heavy LRU-based cache-related overhead, which limits its end-to-end throughput.
In contrast, by employing a \algoname caching algorithm together with speculative sparse prefetching, \sysname achieves an average cache hit ratio of 79.22\% while keeping cache-related overhead lightweight, thereby enabling higher throughput with a much smaller on-GPU cache footprint.

\begin{table}[t!]
\caption{Inference accuracy results.}
\centering
\resizebox{1\linewidth}{!}{   
\begin{tabular}{l|c|cccc}
\toprule
\multicolumn{1}{c|}{\multirow{2}{*}{\textbf{Methods}}}
  & \multirow{2}{*}{\textbf{LongBench}}
  & \multicolumn{4}{c}{\textbf{RULER}} \\ \cline{3-6}
\multicolumn{1}{c|}{}
  &
  & \textbf{16K} & \textbf{32K} & \textbf{64K} & \textbf{128K} \\
\toprule
Qwen2.5-14B              & 53.34 & 94.35 & 94.48 & 92.29 & 88.85 \\
\midrule
RetroInfer               & 54.71 & 94.73 & 94.41 & 92.37 & 89.49 \\
InfiniGen                & 52.74 & 92.78 & 92.19 & 89.74 & 85.26 \\
HATA Algo                     & 53.19 & 94.06 & 94.13 & 92.24 & 88.47 \\
\sysname + HATA          & 53.15 & 94.36 & 94.22 & 92.35 & 88.94 \\
\toprule
Llama3-8B                & 41.06 & 86.07 & 80.80 & 76.26 & 72.96 \\
\midrule
RetroInfer               & 41.11 & 86.36 & 80.64 & 76.26 & 72.73 \\
InfiniGen                & 40.22 & 79.70 & 76.76 & 72.96 & 69.37 \\
HATA Algo                    & 41.58 & 86.56 & 80.94 & 76.28 & 73.87 \\
\sysname + HATA          & 41.15 & 85.65 & 80.70 & 75.94 & 73.44 \\
\bottomrule
\end{tabular}
}
\label{tab:acc}
\end{table}

\subsection{Other Inference Metrics}
\label{sec:eval:metrics}
\noindent\textbf{Accuracy.} 
Table~\ref{tab:acc} reports the average inference accuracy of different methods on LongBench and RULER. 
Since \sysname adopts HATA as its underlying \topk attention algorithm, we additionally include the original HATA \topk algorithm as a baseline to isolate the impact of \algoname and the speculative sparse prefetching strategy on accuracy.
Across both benchmarks, \sysname closely matches the original full-attention models, with a maximum average drop of only 0.42, and it consistently matches or slightly outperforms the HATA algorithm.
On RULER, accuracy remains stable as the sequence length increases, indicating robustness on long-context tasks.
For both Qwen2.5-14B and Llama3-8B, \sysname attains accuracy comparable to the best baselines while avoiding the degradation exhibited by InfiniGen.
Overall, \sysname achieves efficient KVCache offloading without compromising inference quality.

\begin{figure}[t!]
    \centering
    \includegraphics[width=\columnwidth]{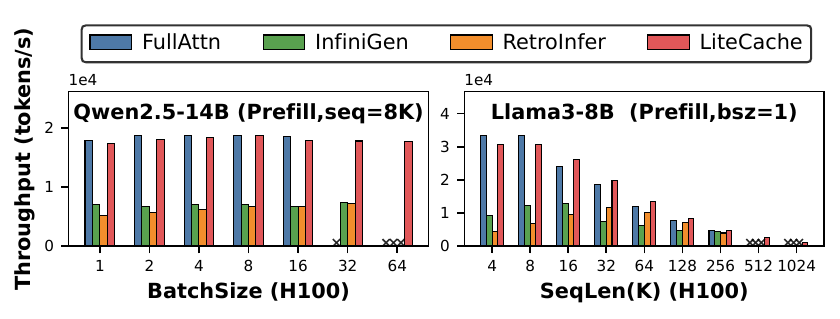}
    \caption{End-to-end prefill throughput across varying batch sizes and sequence lengths on H100.}
    \label{fig:throughput:prefill}
\end{figure}

\noindent\textbf{Prefill performance.} 
Figure~\ref{fig:throughput:prefill} reports the end-to-end prefill throughput on H100. 
Across both models, with various batch sizes and sequence lengths, \sysname closely matches FullAttn, whereas InfiniGen and RetroInfer are consistently slower. This conclusion holds for A40 and more batch size / sequence length configurations, which we omit for brevity.
In prefill, \sysname adds lightweight metadata construction and asynchronous KVCache offloading, whose costs are negligible or largely hidden by computation, so its throughput remains close to FullAttn.
By contrast, RetroInfer spends substantial time on on-the-fly cache construction, and InfiniGen
fails to overlap offloading data transfers with computation in the prefill phase,
resulting in lower prefill throughput.

\begin{figure}[t]
    \centering
    \includegraphics[width=\columnwidth]{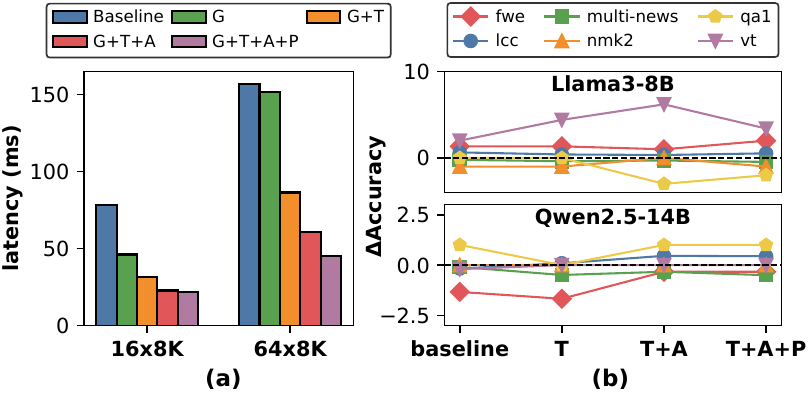}
    \caption{Ablation study of \sysname's optimizations. \textbf{(a) Latency ablation.} Evaluated on Qwen2.5-15B. \textbf{(b) Accuracy ablation.} Evaluated on LongBench (lcc, multi-news) and RULER-128K (nmk2, fwe, qa1, vt).
    }
    \label{fig:eval:ablation}
\end{figure}

\subsection{Ablation Study}
\label{sec:eval:new_ablation}

Figure~\ref{fig:eval:ablation} presents an ablation study of the optimizations applied in \sysname, which include: \textbf{G} (CUDA \textbf{G}raph), \textbf{T} (adaptive \textbf{T}hreshold configuration in Section~\ref{sec:threshold_tuning}), \textbf{A} (intra-GQA \textbf{A}ggregation in Section~\ref{sec:gqa_aggregation}), and \textbf{R} (\textbf{R}esident caching in Section~\ref{sec:outlier_handling}). In addition, \textbf{Baseline} is a simple implementation without these optimizations, which only adopts speculative sparse prefetching and \algoname with a fixed reuse threshold for all the KV heads.


First, in Figure~\ref{fig:eval:ablation}~(a), we report decoding-step latency on an H100 GPU with Qwen2.5-14B under batch sizes of 16 and 64 and sequence lengths of 8K, 16$\times$8K, and 64$\times$8K to show the impact of these optimizations at different data scales.
Under both configurations, the \textbf{Baseline} incurs high latency. Enabling \textbf{G} reduces it by 41.1\% at 16$\times$8K but only 3.5\% at 64$\times$8K, since kernel-launch overhead dominates at small scales but becomes relatively minor at large scales.
Second, \textbf{T} reduces latency by 31.2\% and 43.1\% under the two configurations, followed by \textbf{A} with reductions of 28.3\% and 29.8\%. These gains come from both optimizations improving \algoname’s cache hit rate, thereby reducing PCIe transfer overhead.
Finally, \textbf{R} reduces latency by 3.4\% at 16$\times$8K and 25.2\% at 64$\times$8K, with a stronger effect at larger scales as transfer and computation latency grow. At 16$\times$8K, computation can still hide the PCIe transfer cost of cache-missed KV data via prefetching, but as data volume increases, transfer latency starts to dominate and prefetching can no longer fully mask it. In this regime, handling miss-induced, non-prefetchable outliers via \textbf{R} becomes crucial for maintaining the throughput stability of \sysname.

Next, in Figure~\ref{fig:eval:ablation}~(b), we demonstrate the impact of all optimizations except \textbf{G} on inference accuracy across two models. We use the original HATA algorithm as the baseline and evaluate the change in \sysname's inference accuracy after enabling each optimization (denoted as $\Delta$Accuracy, where positive/negative values indicate improvements/degradations). Although \sysname exhibits accuracy loss on some downstream task benchmarks, most of these are negligible ($\leq$ 1). Furthermore, as described in Section~\ref{sec:eval:metrics}, the average inference accuracy of \sysname remains on par with both HATA and FullAttn. Given the significant performance improvements achieved by \sysname, this trade-off is acceptable.

\section{Related Work}

\para{KVCache offloading.} 
InfLLM~\cite{xiao2024infllm}, MagicPiG~\cite{chen2025magicpig} and ShadowKV~\cite{sun2024shadowkv} focus on improving \topk attention algorithms rather than addressing system-level challenges. 
PQCache~\cite{zhang2025pqcache} and RetroInfer~\cite{chen2025retroinfer} mitigate PCIe transfer volumes through GPU caching, yet they fail to account for the cache management overhead incurred by the CPU. Prefetching approaches, including FlexGen~\cite{sheng2023flexgen} and InfiniGen~\cite{lee2024infinigen}, help alleviate PCIe data transfer latency but face limitations in handling large-scale tasks or entail performance trade-offs. Overall, these methods tend to overlook CPU-side burdens and complexity of implementation. Unlike these,  Mooncake~\cite{qin2025mooncake}, LMCache~\cite{cheng2025lmcache} and CachedAttention~\cite{gao2024cost} optimize only the prefill stage by leveraging distributed or hybrid memory, and are therefore orthogonal to our work.

\para{Kernel launch optimization techniques} have been extensively studied to improve GPU execution efficiency by reducing CPU-side scheduling and launch overhead~\cite{dao2024flashattention, fusion, cuda-graph, compile-time}.
Kernel fusion~\cite{dao2024flashattention, fusion} reduces kernel launch overhead by merging multiple fine-grained GPU operations into a single kernel, allowing intermediate data to be reused in registers or shared memory and reducing redundant global memory accesses.
CUDA Graph~\cite{cuda-graph} further mitigates CPU-side scheduling overhead by replaying static GPU execution graphs, amortizing kernel launch costs across iterations.
Prior work has shown the effectiveness of CUDA Graph for LLM inference during the decoding phase, where involves many short-lived kernels with low compute intensity~\cite{sglang-cuda-graph, vllm-cuda-graph}.
As GPU architectures evolve from A100 to H100 and B200, GPU compute and memory throughput scale far faster than CPU-side scheduling. 
This widening gap further motivates the use of CUDA Graph execution to mitigate CPU-induced bottlenecks.


\section{Conclusion}

\noindent \sysname is an efficient KVCache subsystem for improving LLM inference under GPU HBM constraints.  
Leveraging query similarities, it eliminates cache-related bottlenecks and enables a fully overlapped computation–transfer pipeline. \sysname improves inference throughput by 67.0–224.2\% on H100 and 10.7–60.5\% on A40 compared to strong baselines, while preserving inference accuracy with popular LLMs.





\bibliographystyle{plain}
\bibliography{refs}

@article{vaswani2017attention,
  title={Attention is all you need},
  author={Vaswani, Ashish and Shazeer, Noam and Parmar, Niki and Uszkoreit, Jakob and Jones, Llion and Gomez, Aidan N and Kaiser, {\L}ukasz and Polosukhin, Illia},
  journal={Advances in neural information processing systems},
  volume={30},
  year={2017}
}

@article{touvron2023llama,
  title={Llama 2: Open foundation and fine-tuned chat models},
  author={Touvron, Hugo and Martin, Louis and Stone, Kevin and Albert, Peter and Almahairi, Amjad and Babaei, Yasmine and Bashlykov, Nikolay and Batra, Soumya and Bhargava, Prajjwal and Bhosale, Shruti and others},
  journal={arXiv preprint arXiv:2307.09288},
  year={2023}
}

@Misc{llama3.1,
    title = {Introducing Llama 3.1: Our most capable models to date},
    author={MetaAI},
    year = {2024},
    howpublished = {\url{https://ai.meta.com/blog/meta-llama-3-1/}},
    note = "accessed, Nov. 2025",
}

@Misc{qwen2.5-1m,
    title = {Qwen2.5-1M: Deploy Your Own Qwen with Context Length up to 1M Tokens},
    author={QwenTeam},
    year = {2025},
    howpublished = {\url{https://qwenlm.github.io/blog/qwen2.5-1m/}},
    note = "accessed, Nov. 2025",
}

@article{sun2024shadowkv,
  title={ShadowKV: KV Cache in Shadows for High-Throughput Long-Context LLM Inference},
  author={Sun, Hanshi and Chang, Li-Wen and Bao, Wenlei and Zheng, Size and Zheng, Ningxin and Liu, Xin and Dong, Harry and Chi, Yuejie and Chen, Beidi},
  journal={arXiv preprint arXiv:2410.21465},
  year={2024}
}

@Misc{gdrcopy,
    title = {Magnum IO GDRCopy: Enable faster memory transfers between CPU and GPU with GDRCopy},
    year={2021},
    author={NVIDIA},
    howpublished = {\url{https://developer.nvidia.com/gdrcopy}},
    note = "accessed, Dec. 2025",
}

@article{qin2024llmfornlp,
  title={Large language models meet nlp: A survey},
  author={Qin, Libo and Chen, Qiguang and Feng, Xiachong and Wu, Yang and Zhang, Yongheng and Li, Yinghui and Li, Min and Che, Wanxiang and Yu, Philip S},
  journal={arXiv preprint arXiv:2405.12819},
  year={2024}
}

@article{jiang2024llmforcoding,
  title={A survey on large language models for code generation},
  author={Jiang, Juyong and Wang, Fan and Shen, Jiasi and Kim, Sungju and Kim, Sunghun},
  journal={arXiv preprint arXiv:2406.00515},
  year={2024}
}

@article{yin2024llmformultimodal,
  title={A survey on multimodal large language models},
  author={Yin, Shukang and Fu, Chaoyou and Zhao, Sirui and Li, Ke and Sun, Xing and Xu, Tong and Chen, Enhong},
  journal={National Science Review},
  volume={11},
  number={12},
  pages={nwae403},
  year={2024},
  publisher={Oxford University Press}
}

@Misc{avx,
    title = {AVX-512 Instructions},
    author={Intel},
    year = {2017},
    howpublished = {\url{https://www.intel.com/content/www/us/en/developer/articles/technical/intel-avx-512-instructions.html}},
    note = "accessed, Dec. 2025",
}

@Misc{sglang-cuda-graph,
    title = {DeepSeek-V3 + SGLang: Inference Optimization},
    author={Rodrigo Garcia Casasola},
    year = {2025},
    howpublished = {\url{https://verda.com/blog/deepseek-v3-sglang-inference-optimization}},
    note = "accessed, Dec. 2025",
}

@Misc{vllm-cuda-graph,
    title = {CUDA Graphs},
    author={vLLM},
    year = {2025},
    howpublished = {\url{https://docs.vllm.ai/en/stable/design/cuda_graphs/}},
    note = "accessed, Dec. 2025",
}

@article{fusion,
  title={Optimizing CUDA code by kernel fusion: application on BLAS},
  author={Filipovi{\v{c}}, Ji{\v{r}}{\'\i} and Madzin, Mat{\'u}{\v{s}} and Fousek, Jan and Matyska, Lud{\v{e}}k},
  journal={The Journal of Supercomputing},
  volume={71},
  number={10},
  pages={3934--3957},
  year={2015},
  publisher={Springer}
}

@inproceedings{compile-time,
  title={Chimera: An analytical optimizing framework for effective compute-intensive operators fusion},
  author={Zheng, Size and Chen, Siyuan and Song, Peidi and Chen, Renze and Li, Xiuhong and Yan, Shengen and Lin, Dahua and Leng, Jingwen and Liang, Yun},
  booktitle={2023 IEEE International Symposium on High-Performance Computer Architecture (HPCA)},
  pages={1113--1126},
  year={2023},
  organization={IEEE}
}

@misc{kwon2023efficientmemorymanagementlarge,
      title={Efficient Memory Management for Large Language Model Serving with PagedAttention}, 
      author={Woosuk Kwon and Zhuohan Li and Siyuan Zhuang and Ying Sheng and Lianmin Zheng and Cody Hao Yu and Joseph E. Gonzalez and Hao Zhang and Ion Stoica},
      year={2023},
      eprint={2309.06180},
      archivePrefix={arXiv},
      primaryClass={cs.LG},
      url={https://arxiv.org/abs/2309.06180}, 
}

@Misc{PCIe-Bar,
    title = {PCIe Base Address Registers},
    author={AMD},
    year = {2023},
    howpublished = {\url{https://docs.amd.com/r/en-US/pg055-axi-bridge-pcie/PCIe-Base-Address-Registers}},
    note = "accessed, Dec. 2025",
}

@Misc{cpu-cache-prefetch,
    title = {Cache prefetching},
    author={ArmDeveloper},
    year = {2025},
    howpublished = {\url{https://learn.arm.com/learning-paths/cross-platform/memory-latency/latency-and-cache-prefetching/}},
    note = "accessed, Dec. 2025",
}

@Misc{cuda-graph,
    title = {CUDA Graph},
    author={CUDA Developer},
    year = {2025},
    howpublished = {\url{https://docs.nvidia.com/cuda/cuda-programming-guide/04-special-topics/cuda-graphs.html}},
    note = "accessed, Dec. 2025",
}

@inproceedings{lee2024infinigen,
  title={InfiniGen: Efficient generative inference of large language models with dynamic KV cache management},
  author={Lee, Wonbeom and Lee, Jungi and Seo, Junghwan and Sim, Jaewoong},
  booktitle={18th USENIX Symposium on Operating Systems Design and Implementation (OSDI 24)},
  pages={155--172},
  year={2024}
}

@misc{brown2020languagemodelsfewshotlearners,
      title={Language Models are Few-Shot Learners}, 
      author={Tom B. Brown and Benjamin Mann and Nick Ryder and Melanie Subbiah and Jared Kaplan and Prafulla Dhariwal and Arvind Neelakantan and Pranav Shyam and Girish Sastry and Amanda Askell and Sandhini Agarwal and Ariel Herbert-Voss and Gretchen Krueger and Tom Henighan and Rewon Child and Aditya Ramesh and Daniel M. Ziegler and Jeffrey Wu and Clemens Winter and Christopher Hesse and Mark Chen and Eric Sigler and Mateusz Litwin and Scott Gray and Benjamin Chess and Jack Clark and Christopher Berner and Sam McCandlish and Alec Radford and Ilya Sutskever and Dario Amodei},
      year={2020},
      eprint={2005.14165},
      archivePrefix={arXiv},
      primaryClass={cs.CL},
      url={https://arxiv.org/abs/2005.14165}, 
}

@Misc{nvidia_uva,
  title={CUDA C++ Programming Guide},
  author={{NVIDIA}},
  howpublished = {\url{https://docs.nvidia.com/cuda/cuda-c-programming-guide/#unified-virtual-address-space}},
  year = {2025},
  note = "accessed, Dec. 2025",
}

@inproceedings{sheng2023flexgen,
  title={Flexgen: High-throughput generative inference of large language models with a single gpu},
  author={Sheng, Ying and Zheng, Lianmin and Yuan, Binhang and Li, Zhuohan and Ryabinin, Max and Chen, Beidi and Liang, Percy and R{\'e}, Christopher and Stoica, Ion and Zhang, Ce},
  booktitle={International Conference on Machine Learning},
  pages={31094--31116},
  year={2023},
  organization={PMLR}
}

@article{zhang2025pqcache,
  title={Pqcache: Product quantization-based kvcache for long context llm inference},
  author={Zhang, Hailin and Ji, Xiaodong and Chen, Yilin and Fu, Fangcheng and Miao, Xupeng and Nie, Xiaonan and Chen, Weipeng and Cui, Bin},
  journal={Proceedings of the ACM on Management of Data},
  volume={3},
  number={3},
  pages={1--30},
  year={2025},
  publisher={ACM New York, NY, USA}
}

@article{xiao2024infllm,
  title={Infllm: Training-free long-context extrapolation for llms with an efficient context memory},
  author={Xiao, Chaojun and Zhang, Pengle and Han, Xu and Xiao, Guangxuan and Lin, Yankai and Zhang, Zhengyan and Liu, Zhiyuan and Sun, Maosong},
  journal={Advances in Neural Information Processing Systems},
  volume={37},
  pages={119638--119661},
  year={2024}
}

@inproceedings{
chen2025magicpig,
    title={Magic{PIG}: {LSH} Sampling for Efficient {LLM} Generation},
    author={Zhuoming Chen and Ranajoy Sadhukhan and Zihao Ye and Yang Zhou and Jianyu Zhang and Niklas Nolte and Yuandong Tian and Matthijs Douze and Leon Bottou and Zhihao Jia and Beidi Chen},
    booktitle={The Thirteenth International Conference on Learning Representations},
    year={2025},
    url={https://openreview.net/forum?id=ALzTQUgW8a}
}

@article{chen2025retroinfer,
  title={RetroInfer: A Vector-Storage Approach for Scalable Long-Context LLM Inference},
  author={Chen, Yaoqi and Zhang, Jinkai and Lu, Baotong and Zhang, Qianxi and Zhang, Chengruidong and Luo, Jingjia and Liu, Di and Jiang, Huiqiang and Chen, Qi and Liu, Jing and others},
  journal={arXiv preprint arXiv:2505.02922},
  year={2025}
}

@inproceedings{tang2024quest,
author = {Tang, Jiaming and Zhao, Yilong and Zhu, Kan and Xiao, Guangxuan and Kasikci, Baris and Han, Song},
title = {QUEST: query-aware sparsity for efficient long-context LLM inference},
year = {2024},
publisher = {JMLR.org},
booktitle = {Proceedings of the 41st International Conference on Machine Learning},
articleno = {1955},
numpages = {11},
location = {Vienna, Austria},
series = {ICML'24}
}

@article{singhania2024loki,
  title={Loki: Low-rank keys for efficient sparse attention},
  author={Singhania, Prajwal and Singh, Siddharth and He, Shwai and Feizi, Soheil and Bhatele, Abhinav},
  journal={Advances in Neural Information Processing Systems},
  volume={37},
  pages={16692--16723},
  year={2024}
}

@inproceedings{gong2025hata,
    title = "{HATA}: Trainable and Hardware-Efficient Hash-Aware Top-$k$ Attention for Scalable Large Model Inference",
    author = "Gong, Ping  and
      Yi, Jiawei  and
      Wang, Shengnan  and
      Zhang, Juncheng  and
      Jin, Zewen  and
      Zhou, Ouxiang  and
      Liu, Ruibo  and
      Xu, Guanbin  and
      Bai, Youhui  and
      Ye, Bowen  and
      Yuan, Kun  and
      Yang, Tong  and
      Zhang, Gong  and
      Chen, Renhai  and
      Wu, Feng  and
      Li, Cheng",
    editor = "Che, Wanxiang  and
      Nabende, Joyce  and
      Shutova, Ekaterina  and
      Pilehvar, Mohammad Taher",
    booktitle = "Findings of the Association for Computational Linguistics: ACL 2025",
    month = jul,
    year = "2025",
    address = "Vienna, Austria",
    publisher = "Association for Computational Linguistics",
    url = "https://aclanthology.org/2025.findings-acl.1275/",
    pages = "24856--24871",
    ISBN = "979-8-89176-256-5"
}

@inproceedings{gupta2021topkattn,
    title = "Memory-efficient Transformers via Top-k Attention",
    author = "Gupta, Ankit  and
      Dar, Guy  and
      Goodman, Shaya  and
      Ciprut, David  and
      Berant, Jonathan",
    editor = "Moosavi, Nafise Sadat  and
      Gurevych, Iryna  and
      Fan, Angela  and
      Wolf, Thomas  and
      Hou, Yufang  and
      Marasovi{\'c}, Ana  and
      Ravi, Sujith",
    booktitle = "Proceedings of the Second Workshop on Simple and Efficient Natural Language Processing",
    month = nov,
    year = "2021",
    address = "Virtual",
    publisher = "Association for Computational Linguistics",
    url = "https://aclanthology.org/2021.sustainlp-1.5/",
    doi = "10.18653/v1/2021.sustainlp-1.5",
    pages = "39--52"
}

@inproceedings{wolf-etal-2020-transformers,
    title = "Transformers: State-of-the-Art Natural Language Processing",
    author = "Thomas Wolf and Lysandre Debut and Victor Sanh and Julien Chaumond and Clement Delangue and Anthony Moi and Pierric Cistac and Tim Rault and Rémi Louf and Morgan Funtowicz and Joe Davison and Sam Shleifer and Patrick von Platen and Clara Ma and Yacine Jernite and Julien Plu and Canwen Xu and Teven Le Scao and Sylvain Gugger and Mariama Drame and Quentin Lhoest and Alexander M. Rush",
    booktitle = "Proceedings of the 2020 Conference on Empirical Methods in Natural Language Processing: System Demonstrations",
    month = oct,
    year = "2020",
    address = "Online",
    publisher = "Association for Computational Linguistics",
    url = "https://www.aclweb.org/anthology/2020.emnlp-demos.6",
    pages = "38--45"
}

@inproceedings{
dao2024flashattention,
title={FlashAttention-2: Faster Attention with Better Parallelism and Work Partitioning},
author={Tri Dao},
booktitle={The Twelfth International Conference on Learning Representations},
year={2024},
url={https://openreview.net/forum?id=mZn2Xyh9Ec}
}

@article{ye2025flashinfer,
    title = {FlashInfer: Efficient and Customizable Attention Engine for LLM Inference Serving},
    author = {
      Ye, Zihao and
      Chen, Lequn and
      Lai, Ruihang and
      Lin, Wuwei and
      Zhang, Yineng and
      Wang, Stephanie and
      Chen, Tianqi and
      Kasikci, Baris and
      Grover, Vinod and
      Krishnamurthy, Arvind and
      Ceze, Luis
    },
    journal = {arXiv preprint arXiv:2501.01005},
    year = {2025},
    url = {https://arxiv.org/abs/2501.01005}
}

@inproceedings{ribar2023sparq,
author = {Ribar, Luka and Chelombiev, Ivan and Hudlass-Galley, Luke and Blake, Charlie and Luschi, Carlo and Orr, Douglas},
title = {SparQ attention: bandwidth-efficient LLM inference},
year = {2024},
publisher = {JMLR.org},
booktitle = {Proceedings of the 41st International Conference on Machine Learning},
articleno = {1731},
numpages = {26},
location = {Vienna, Austria},
series = {ICML'24}
}

@article{bairi2024codeplan,
  title={Codeplan: Repository-level coding using llms and planning},
  author={Bairi, Ramakrishna and Sonwane, Atharv and Kanade, Aditya and C, Vageesh D and Iyer, Arun and Parthasarathy, Suresh and Rajamani, Sriram and Ashok, Balasubramanyan and Shet, Shashank},
  journal={Proceedings of the ACM on Software Engineering},
  volume={1},
  number={FSE},
  pages={675--698},
  year={2024},
  publisher={ACM New York, NY, USA}
}

@inproceedings{gao2024multiturnconversation,
  title={Cost-Efficient large language model serving for multi-turn conversations with CachedAttention},
  author={Gao, Bin and He, Zhuomin and Sharma, Puru and Kang, Qingxuan and Jevdjic, Djordje and Deng, Junbo and Yang, Xingkun and Yu, Zhou and Zuo, Pengfei},
  booktitle={2024 USENIX Annual Technical Conference (USENIX ATC 24)},
  pages={111--126},
  year={2024}
}

@article{koh2022documentsum,
  title={An empirical survey on long document summarization: Datasets, models, and metrics},
  author={Koh, Huan Yee and Ju, Jiaxin and Liu, Ming and Pan, Shirui},
  journal={ACM computing surveys},
  volume={55},
  number={8},
  pages={1--35},
  year={2022},
  publisher={ACM New York, NY}
}

@inproceedings{
hsieh2024ruler,
title={{RULER}: What{\textquoteright}s the Real Context Size of Your Long-Context Language Models?},
author={Cheng-Ping Hsieh and Simeng Sun and Samuel Kriman and Shantanu Acharya and Dima Rekesh and Fei Jia and Boris Ginsburg},
booktitle={First Conference on Language Modeling},
year={2024},
url={https://openreview.net/forum?id=kIoBbc76Sy}
}

@misc{gradientlongcontextllama3,
  title={Llama 3 Gradient: A series of long context models},
  author={Leonid Pekelis and Michael Feil and Forrest Moret and Mark Huang and Tiffany Peng},
  year={2024},
  url = {https://gradient.ai/blog/scaling-rotational-embeddings-for-long-context-language-models},
  doi = { 10.57967/hf/3372 },
}

@inproceedings{
    xiao2024streamingllm,
    title={Efficient Streaming Language Models with Attention Sinks},
    author={Guangxuan Xiao and Yuandong Tian and Beidi Chen and Song Han and Mike Lewis},
    booktitle={The Twelfth International Conference on Learning Representations},
    year={2024},
    url={https://openreview.net/forum?id=NG7sS51zVF}
}

@inproceedings{
xiao2025duoattention,
title={DuoAttention: Efficient Long-Context {LLM} Inference with Retrieval and Streaming Heads},
author={Guangxuan Xiao and Jiaming Tang and Jingwei Zuo and junxian guo and Shang Yang and Haotian Tang and Yao Fu and Song Han},
booktitle={The Thirteenth International Conference on Learning Representations},
year={2025},
url={https://openreview.net/forum?id=cFu7ze7xUm}
}

@inproceedings{bai-etal-2024-longbench,
    title = "{L}ong{B}ench: A Bilingual, Multitask Benchmark for Long Context Understanding",
    author = "Bai, Yushi  and
      Lv, Xin  and
      Zhang, Jiajie  and
      Lyu, Hongchang  and
      Tang, Jiankai  and
      Huang, Zhidian  and
      Du, Zhengxiao  and
      Liu, Xiao  and
      Zeng, Aohan  and
      Hou, Lei  and
      Dong, Yuxiao  and
      Tang, Jie  and
      Li, Juanzi",
    editor = "Ku, Lun-Wei  and
      Martins, Andre  and
      Srikumar, Vivek",
    booktitle = "Proceedings of the 62nd Annual Meeting of the Association for Computational Linguistics (Volume 1: Long Papers)",
    month = aug,
    year = "2024",
    address = "Bangkok, Thailand",
    publisher = "Association for Computational Linguistics",
    url = "https://aclanthology.org/2024.acl-long.172/",
    doi = "10.18653/v1/2024.acl-long.172",
    pages = "3119--3137"
}

@article{o1993lru,
  title={The LRU-K page replacement algorithm for database disk buffering},
  author={O'neil, Elizabeth J and O'neil, Patrick E and Weikum, Gerhard},
  journal={Acm Sigmod Record},
  volume={22},
  number={2},
  pages={297--306},
  year={1993},
  publisher={ACM New York, NY, USA}
}

@inproceedings{megiddo2003arc,
  title={ARC: A Self-Tuning, low overhead replacement cache},
  author={Megiddo, Nimrod and Modha, Dharmendra S},
  booktitle={2nd USENIX Conference on File and Storage Technologies (FAST 03)},
  year={2003}
}

@inproceedings{robinson1990data,
  title={Data cache management using frequency-based replacement},
  author={Robinson, John T and Devarakonda, Murthy V},
  booktitle={Proceedings of the 1990 ACM SIGMETRICS conference on Measurement and modeling of computer systems},
  pages={134--142},
  year={1990}
}

@misc{glm2024chatglmfamilylargelanguage,
      title={ChatGLM: A Family of Large Language Models from GLM-130B to GLM-4 All Tools}, 
      author={Team GLM and Aohan Zeng and Bin Xu and Bowen Wang and Chenhui Zhang and others},
      year={2024},
      eprint={2406.12793},
      archivePrefix={arXiv},
      primaryClass={cs.CL},
      url={https://arxiv.org/abs/2406.12793}, 
}

@misc{deepseekai2024deepseekv2strongeconomicalefficient,
      title={DeepSeek-V2: A Strong, Economical, and Efficient Mixture-of-Experts Language Model}, 
      author={DeepSeek-AI and Aixin Liu and Bei Feng and Bin Wang and others},
      year={2024},
      eprint={2405.04434},
      archivePrefix={arXiv},
      primaryClass={cs.CL},
      url={https://arxiv.org/abs/2405.04434}, 
}

@misc{openai2025gptoss120bgptoss20bmodel,
      title={gpt-oss-120b and gpt-oss-20b Model Card}, 
      author={OpenAI and Sandhini Agarwal and Lama Ahmad and Jason Ai and Sam Altman and others},
      year={2025},
      eprint={2508.10925},
      archivePrefix={arXiv},
      primaryClass={cs.CL},
      url={https://arxiv.org/abs/2508.10925}, 
}

@misc{gao2025seerattentionlearningintrinsicsparse,
      title={SeerAttention: Learning Intrinsic Sparse Attention in Your LLMs}, 
      author={Yizhao Gao and Zhichen Zeng and Dayou Du and Shijie Cao and Peiyuan Zhou and Jiaxing Qi and Junjie Lai and Hayden Kwok-Hay So and Ting Cao and Fan Yang and Mao Yang},
      year={2025},
      eprint={2410.13276},
      archivePrefix={arXiv},
      primaryClass={cs.CL},
      url={https://arxiv.org/abs/2410.13276}, 
}

@article{su2024roformer,
  title={Roformer: Enhanced transformer with rotary position embedding},
  author={Su, Jianlin and Ahmed, Murtadha and Lu, Yu and Pan, Shengfeng and Bo, Wen and Liu, Yunfeng},
  journal={Neurocomputing},
  volume={568},
  pages={127063},
  year={2024},
  publisher={Elsevier}
}

@misc{LiteCache2025,
  author       = {},
  title        = {Source Code of LiteCache},
  howpublished = {\url{https://anonymous.4open.science/r/LiteCache-888D}},
  year         = {2025},
  note = "accessed, Dec. 2025",
}

@misc{H100Whitepaper2025,
  author       = {NVIDIA},
  title        = {NVIDIA H100 GPU Whitepaper},
  howpublished = {\url{https://resources.nvidia.com/en-us-hopper-architecture/nvidia-h100-tensor-c}},
  year         = {2025},
  note = "accessed, Dec. 2025",
}

@inproceedings{qin2025mooncake,
  title={Mooncake: Trading more storage for less computation—a KVCache-centric architecture for serving LLM chatbot},
  author={Qin, Ruoyu and Li, Zheming and He, Weiran and Cui, Jialei and Ren, Feng and Zhang, Mingxing and Wu, Yongwei and Zheng, Weimin and Xu, Xinran},
  booktitle={23rd USENIX Conference on File and Storage Technologies (FAST 25)},
  pages={155--170},
  year={2025}
}

@article{cheng2025lmcache,
  title={Lmcache: An efficient KV cache layer for enterprise-scale LLM inference},
  author={Cheng, Yihua and Liu, Yuhan and Yao, Jiayi and An, Yuwei and Chen, Xiaokun and Feng, Shaoting and Huang, Yuyang and Shen, Samuel and Du, Kuntai and Jiang, Junchen},
  journal={arXiv preprint arXiv:2510.09665},
  year={2025}
}

@inproceedings{gao2024cost,
  title={Cost-Efficient large language model serving for multi-turn conversations with CachedAttention},
  author={Gao, Bin and He, Zhuomin and Sharma, Puru and Kang, Qingxuan and Jevdjic, Djordje and Deng, Junbo and Yang, Xingkun and Yu, Zhou and Zuo, Pengfei},
  booktitle={2024 USENIX Annual Technical Conference (USENIX ATC 24)},
  pages={111--126},
  year={2024}
}


\end{document}